\documentclass[11pt]{article}

\usepackage[preprint]{acl}

\usepackage{times}
\usepackage{latexsym}
\usepackage{amssymb}    

\usepackage[T1]{fontenc}

\usepackage[utf8]{inputenc}

\usepackage{microtype}

\usepackage{inconsolata}

\usepackage{graphicx}

\usepackage{booktabs}       
\usepackage{xcolor}         
\usepackage[table]{xcolor} 
\usepackage{graphicx}
\graphicspath{%
  {./}%
  {figures/}%
  {sections/appendix/appendix_skill_overview/figures/}%
  {sections/appendix/appendix_inference_loading/figures/}%
  {sections/appendix/appendix_training_supervision/figures/}%
}
\usepackage{colortbl}
\usepackage{enumitem}
\usepackage{xspace}
\usepackage{pifont}
\usepackage{tabularx}
\usepackage[most]{tcolorbox}
\usepackage{listings}
\usepackage{subcaption}
\usepackage[most]{tcolorbox}
\usepackage{enumitem}

\newcommand{\ours}{\textsc{VisualSkill}\xspace}

\title{\ours: Multimodal Skills for Computer-Use Agents}

\author{
  \textbf{Ziyan Jiang\textsuperscript{1,*}},
  \textbf{Li An\textsuperscript{1,*}},
  \textbf{Yujian Liu\textsuperscript{1}},
  \textbf{Jiabao Ji\textsuperscript{1}},
  \textbf{Qiucheng Wu\textsuperscript{1}},
\\
  \textbf{Jacob Andreas\textsuperscript{2,\textdagger}},
  \textbf{Yang Zhang\textsuperscript{3,\textdagger}},
  \textbf{Shiyu Chang\textsuperscript{1,\textdagger}}
\\
\\
  \textsuperscript{1}UC Santa Barbara,\quad
  \textsuperscript{2}MIT CSAIL,\quad
  \textsuperscript{3}MIT-IBM Watson AI Lab
\\
  \small\textsuperscript{*}Equal contribution.\quad\textsuperscript{\textdagger}Equal advising.\quad Correspondence to: \texttt{\{ziyanjiang, li\_an\}@ucsb.edu}
}

\begin{document}
\maketitle

\begin{abstract}
Computer-use agents (CUAs) approach human-level performance on standardised benchmarks but still struggle on long-horizon tasks and unseen software. Existing skill libraries address this with reusable skills, but represent the skill artifact as text only, despite the visual nature of GUI interaction. We propose \ours: a hierarchical multimodal skill, tailored to each target application and organised as a central index over per-topic files, which the agent consumes through a \texttt{load\_topic} MCP tool that fetches the relevant topic's text and figures on demand. We construct each skill with a two-stage pipeline that combines authored documentation with live-application UI exploration. On two CUA benchmarks, CUA-World and OSExpert-Eval, a Claude Code CLI agent backed by Claude Opus~4.6 reaches an average score of $0.456$ with \ours{}, a $\mathbf{+15.3}$ point absolute lift over the no-skill baseline ($0.303$). Against a matched text-only skill that is generated from the same source content and differs from \ours{} only in modality, \ours{} yields a further $\mathbf{+8.3}$ point absolute gain over the matched text-only skill ($0.373$ vs.\ $0.456$), providing direct evidence that retaining visual figures in the skill artifact, rather than verbalizing them away, helps the agent both identify UI elements and verify workflow state after each action. Our code is available at \url{https://github.com/XMHZZ2018/VisualSkills}.
\end{abstract}

\section{Introduction}
\label{sec:intro}

Agent skills recently introduce a modular, filesystem-based abstraction that equips agents with domain-specific expertise on demand. Skills are particularly valuable for computer-use agents (CUAs), which interact with graphical user interfaces through screenshots and keyboard/mouse actions. CUAs have advanced rapidly~\citep{wang2026opencua, agashe2025agent} and now approach human-level performance on standardised desktop benchmarks such as OSWorld~\citep{xie2024osworld, claude_sonnet_46, agent_s3}, yet they still struggle on complex, long-horizon tasks and generalise poorly to unseen UIs and software~\citep{aggarwal2026gym, liu2026osexpert}, largely because they lack the persistent, application-specific procedural knowledge (which menu hides a given command, what dialog appears after a particular click, how a multi-step workflow unfolds) that pre-training does not supply. A skill can supply exactly that knowledge at decision time.

\begin{figure}[t]
    \centering
    \includegraphics[width=\columnwidth]{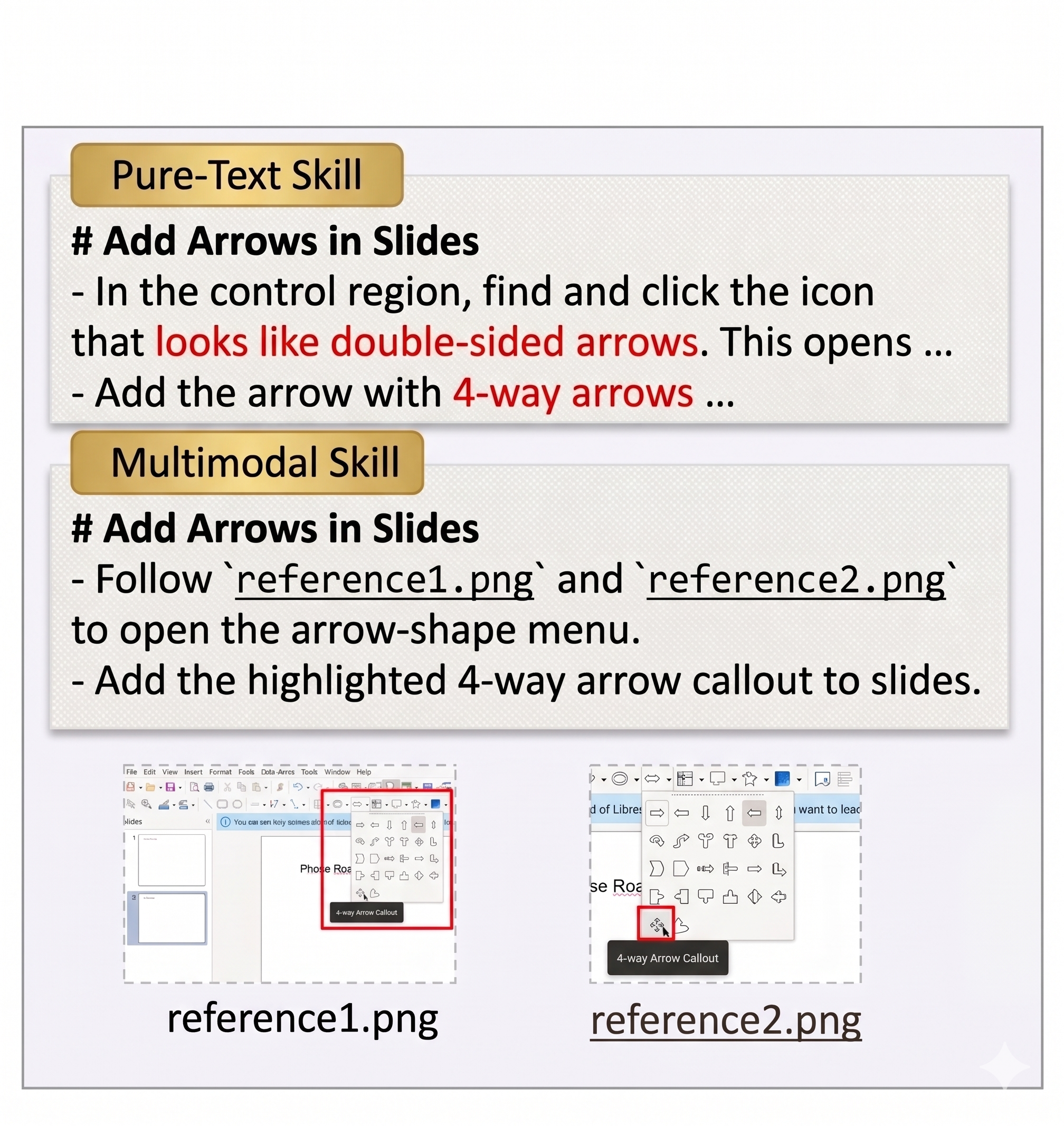}
    \caption{A text-only skill struggles to describe the precise operations and icons to focus on, creating ambiguities that multimodal skills can avoid.}
    \label{fig:motivation_example}
    \vspace{-4mm}
\end{figure}

However, existing skill libraries for CUAs are predominantly text-only, representing the skill artifact as a natural-language intent paired with text-based action specifications and providing no slot for figures or screenshots~\citep{chen2026cua, liu2026osexpert}, which is a poor match for computer-use environments in two ways. \textbf{First}, many UI elements that an agent must act on (icons, layouts, interface states, and spatial relations among widgets) are verbose or ambiguous to describe in text, so the verbal substitutes that enter the skill lose information that the original screenshot carries directly. \textbf{Second}, multi-step workflows require the agent to verify, after each action, that it has reached the expected intermediate UI state. A reference screenshot serves as a direct visual grounding signal for the target UI state and can be matched against the agent’s current observation. A textual description, however, specifies the state only indirectly, making it more difficult to verify whether the agent has reached the intended state or a visually similar but incorrect one. These two limitations motivate \emph{multimodal skills} that retain visual content as figures alongside the textual procedure. As shown in Figure~\ref{fig:motivation_example}, a text-only skill has difficulty describing the precise operations and icons to focus on, creating ambiguity that a multimodal skill can avoid.

We introduce \ours, which constructs multimodal skills in which visual content is retained as figures. \ours produces \emph{one skill per target application}, scoping a skill to all the UI knowledge an agent needs to operate that application end to end. Each such skill is organised \emph{hierarchically}: a central index file lists every topic with a short ``when to use'' description, and points to a per-topic file that holds the text body and figures for that topic. At inference time, the agent reads only the compact index and invokes a \texttt{load\_topic} MCP tool to fetch the text and figures on demand. 

To build such skills, \ours runs a two-stage pipeline in which each stage targets one of the two challenges above. Stage~1 mines authored documentation (PDF guides, HTML manuals) into this hierarchy, with figures from the source documents kept verbatim. Stage~2 enriches the skill by exploring the live application under two regimes: \emph{free exploration}, in which an LLM planner partitions the application's starting page into regions to cover, and \emph{targeted exploration}, in which training-task trajectories pinpoint UI regions where the current skill underperforms. Screenshots collected from both regimes are, after optional post-processing, slotted into the corresponding per-topic skill files.

In summary, we make three contributions. First, we design \ours{} as hierarchical multimodal artefacts with one skill per target application and a central index over per-topic files, which the agent consumes through a \texttt{load\_topic} MCP tool that fetches only the relevant topic's text and figures on demand. Second, we construct such skills with a two-stage pipeline that combines authored documentation (Stage~1) with live-application UI exploration (Stage~2). Third, we evaluate \ours{} against a text-only control skill that is generated jointly from the same source content and differs only in modality, so any performance gap isolates the effect of visual presentation: across $177$ tasks from CUA-World and OSExpert-Eval, using a Claude Opus~4.6 Claude Code CLI agent, \ours{} lifts the unweighted-mean score from $0.303$ (no-skill) to $0.456$ ($+15.3\%$ absolute) and outperforms the text-only control by $+8.3\%$ absolute, while a qualitative analysis shows that the multimodal advantage concentrates on the two failure modes of text-only skills identified above: identifying UI elements (icons, layouts, spatial relations) and verifying intermediate workflow state after each action.

\section{Method}
\label{sec:method}

This section defines what a \ours is, how the agent uses it at inference, and how we construct one for a target desktop application. Section~\ref{sec:method-def} gives the formal definition of \ours. Section~\ref{sec:method-loading} describes the loading mechanism by which the agent navigates the skill at decision time. Section~\ref{sec:method-construct} describes the two-stage pipeline that produces a skill for any application from authored documentation (Stage~1) and from interaction with the live application itself (Stage~2).

\subsection{Skill Definition}
\label{sec:method-def}

A skill in \ours is a structured reference about a single target application: one skill per application, shared by all tasks against that application.

Each skill is organised as a centralised \texttt{skill.md} index plus a list of per-topic guides $\{g_t\}_{t \in \mathcal{T}}$ arranged in a two-layer hierarchy. The \texttt{skill.md} index points to each per-topic guide and tags it with a one-sentence \emph{when to use} description that the agent matches against its current task before loading the guide. Each per-topic guide $g_t = (p_t, F_t)$ pairs a text body $p_t$ with a set of UI figures $F_t$. At inference, the agent reads only the \texttt{skill.md} index up front and loads individual $g_t$ on demand (Section~\ref{sec:method-loading}). See Appendix~\ref{app:skill-overview-structure} for the full structure and an excerpt of \texttt{skill.md}.

For each application, the artifact we deliver is a \ours{} $\mathcal{S}^{\mathrm{mm}}$ that retains the UI figures. To isolate the contribution of the visual modality at evaluation time, we additionally construct a \textbf{text-only control} $\mathcal{S}^{\mathrm{txt}}$ from the same source material: it shares the same hierarchical structure as $\mathcal{S}^{\mathrm{mm}}$ (same \texttt{skill.md} index, same topic set $\mathcal{T}$, same procedural content per topic) and differs only in the per-topic guides --- $\mathcal{S}^{\mathrm{mm}}$ uses $g_t^{\mathrm{mm}} = (p_t^{\mathrm{mm}}, F_t)$ with the UI figures alongside the text body, while $\mathcal{S}^{\mathrm{txt}}$ uses $g_t^{\mathrm{txt}} = (p_t^{\mathrm{txt}}, \emptyset)$ with no figures and a text body that describes the same visual information directly in words. The two text bodies are not word-for-word identical, since each is written in the form most natural to its modality. To guarantee that the procedural content is shared, every per-topic generation step in the pipeline (Section~\ref{sec:method-construct}) is a single LLM call that reads the topic's source text and figures and emits $p_t^{\mathrm{mm}}$ and $p_t^{\mathrm{txt}}$ side by side. Under this construction, any gap in agent performance between $\mathcal{S}^{\mathrm{mm}}$ and $\mathcal{S}^{\mathrm{txt}}$ is attributable to modality alone, not to differences in the underlying content. See Appendix~\ref{app:skill-overview-matched} for an excerpt of $\mathcal{S}^{\mathrm{mm}}$ and the corresponding $\mathcal{S}^{\mathrm{txt}}$ on one topic.

\subsection{Skill Loading at Inference Time}
\label{sec:method-loading}

At inference, the agent does not load the entire skill up front. Its system prompt contains only the \texttt{SKILL.md} index, which lists each topic together with a one-sentence \emph{when to use} description. To retrieve a per-topic guide, the agent calls a single MCP tool, \texttt{load\_topic(t)}, which takes a topic identifier $t \in \mathcal{T}$ and returns the guide $g_t$. At every step, the agent matches its next intended action against the \emph{when to use} lines and calls \texttt{load\_topic} before acting if any of them matches. The tool may be called multiple times across a trajectory as the task moves between UI surfaces.

For the multimodal variant, the returned content interleaves text and image blocks in the order figures are referenced by $p_t$, so each image of $F_t$ is delivered immediately after the sentence that names it. For the text-only control, the same call returns a single text block whose wording already absorbs the visual information. The full system prompt, the \texttt{load\_topic} schema, and a worked tool-call transcript are provided in Appendix~\ref{app:inference-loading}.

We expose skill loading through an MCP tool rather than direct \texttt{Read} access for two reasons. First, the MCP tool delivers each figure inline with its surrounding text in a single tool result, whereas under direct \texttt{Read} each figure incurs a separate call and is skipped in practice. Second, the MCP interface keeps skill content accessible to the agent throughout the trajectory, whereas direct \texttt{Read} tends to be invoked once at the start and not revisited. Section~\ref{sec:exp-loading} verifies these effects empirically: under direct \texttt{Read}, the agent loads $\sim$$10\times$ fewer figures per task and stops consulting the skill within the first $\sim$2\% of the rollout, collapsing the multimodal gain.

\subsection{Two-Stage Skill Construction}
\label{sec:method-construct}

\begin{figure*}[t]
    \centering
    \includegraphics[width=0.95\textwidth]{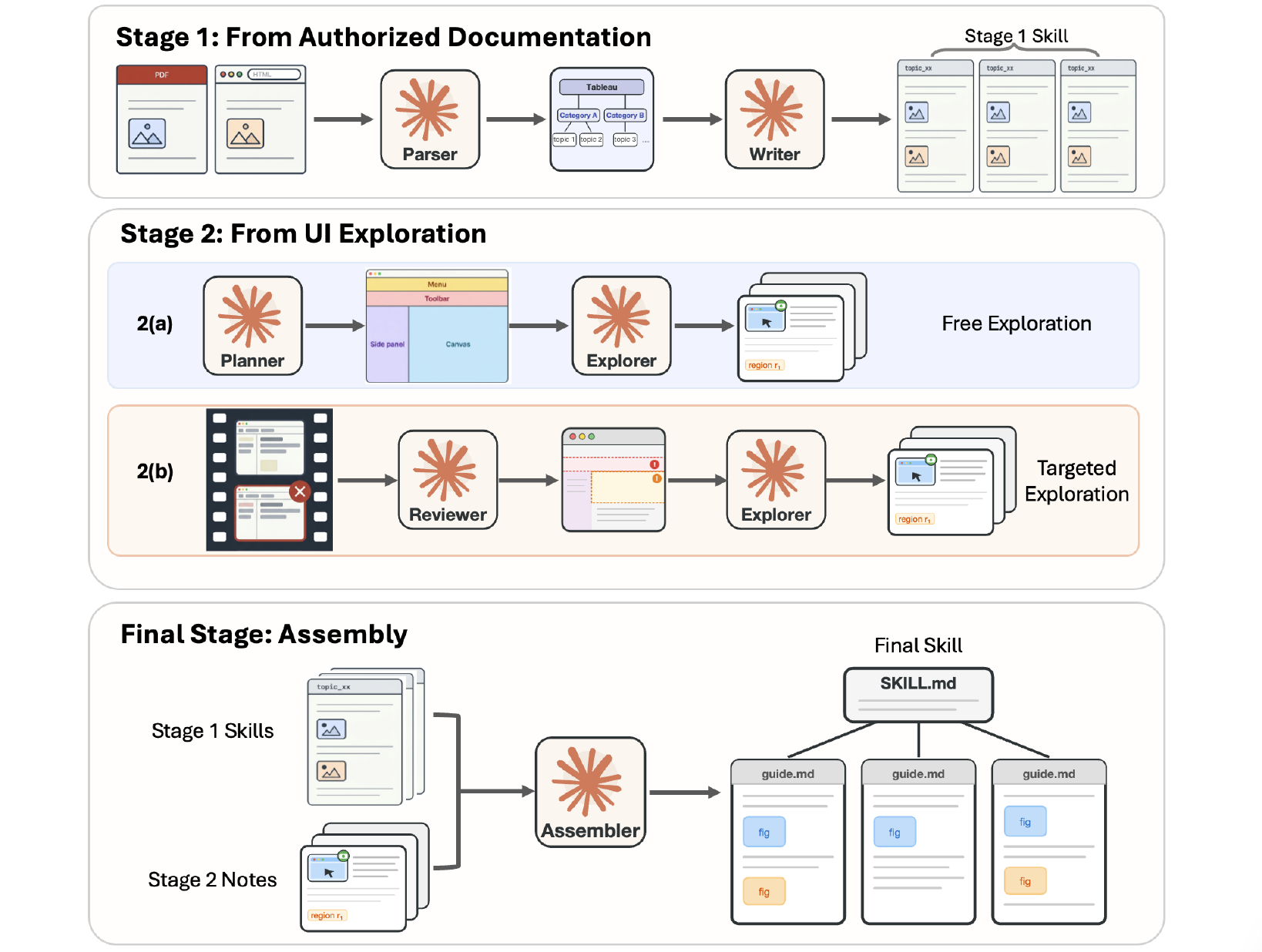}
    \caption{The two-stage \ours{} construction pipeline. \textbf{Stage~1} parses the authored documentation into a topic hierarchy, extracting per-topic text bodies and the vendor-drawn figures shipped with the manual. \textbf{Stage~2} drives the live application with an LLM-controlled explorer in two sub-passes --- a \emph{free explorer} that partitions the idle window and dispatches a worker per region, and a \emph{trajectory-targeted explorer} that re-reads failed rollouts and patches the UI regions agents misread --- and merges captured screenshots and notes into the per-topic guides. The matched text-only skill is generated jointly from the same source, differing only in modality.}
    \label{fig:pipeline-overview}
\end{figure*}

Our pipeline (Figure~\ref{fig:pipeline-overview}) constructs the \ours{} skill $\mathcal{S}^{\mathrm{mm}}$ for an application $\mathcal{A}$ in two stages, generating the text-only control $\mathcal{S}^{\mathrm{txt}}$ alongside it from the same source. \textbf{Stage~1} mines an initial version of the skill from authored documentation, exploiting the fact that mature applications usually ship a structured manual whose table of contents can be reused as the topic set $\mathcal{T}$. \textbf{Stage~2} augments it by driving the live application with an LLM-controlled explorer, capturing UI knowledge that exists only in the running program. At each stage, the same single-call protocol of Section~\ref{sec:method-def} emits both $\mathcal{S}^{\mathrm{mm}}$ and $\mathcal{S}^{\mathrm{txt}}$ from the same input, so $\mathcal{T}$ and the per-topic procedural content are shared by construction; only $F_t$ and the wording of $p_t$ around each figure-slot differ.

\paragraph{Stage 1: from authored documentation.}
Stage~1 takes the application's official user guide $D$ (a PDF or HTML manual in our experiments) and produces the Stage~1 \ours{} skill $\mathcal{S}^{\mathrm{mm}}_1$ in a few LLM-driven steps; the text-only control $\mathcal{S}^{\mathrm{txt}}_1$ is emitted in the same calls. We first parse $D$'s table of contents and reuse it as $\mathcal{T}$: the application's maintainers have already organised the surface for us, and inheriting their structure is cheaper and more faithful than imposing a fresh one. For each topic $t$, we then build the per-topic guide $g_t = (p_t, F_t)$ by locating the pages of $D$ that cover $t$, pulling out every figure on those pages as $F_t$, and invoking the joint-generation call to write $p_t^{\mathrm{mm}}$ and $p_t^{\mathrm{txt}}$ side by side. Because the figures come straight from the official manual, $F_t$ holds vendor-drawn screenshots of the application's UI rather than agent-captured ones. The resulting skill is hierarchically organised, and its coverage is whatever the documentation covers. Appendix~\ref{app:examples-stage1} walks through one Stage~1 topic end-to-end.

\paragraph{Stage 2: from UI exploration.}
The Stage~1 skill is unavoidably incomplete: documentation is often outdated relative to the shipped UI, omits low-frequency dialogs, and rarely captures the visual affordances that an agent must distinguish at click time. Stage~2 closes this gap by driving the live application with a planner and a pool of worker agents. We run two complementary sub-passes whose outputs are merged before reassembly.

\textit{(a) Free explorer.} An Opus-class planner agent $P$ inspects a screenshot of the idle application and nominates a partition of its UI into a set of regions $R = \{r_1, \ldots, r_K\}$ (with $K{=}8$ in our experiments). For each region $r_i$, a Sonnet-class worker agent $W_i$ is spawned in an isolated Docker container running $\mathcal{A}$, instructed to interact with $r_i$, capture cropped screenshots of every interactive element, and emit a structured note $n_i$ summarising the affordances of $r_i$. Workers run in parallel. Appendix~\ref{app:examples-stage2a} walks through one free-explorer region end-to-end.

\textit{(b) Trajectory-targeted explorer.} In parallel, a reviewer agent $V$ is given a held-out set of training-task trajectories $\{\tau_1, \ldots, \tau_N\}$ on which the Stage~1 skill was executed and inspected for failures, where each $\tau_j$ contains the task instruction, the agent's action sequence, the final verifier feedback, and any consultations the agent issued. $V$ produces an additional set of UI regions $R' = V(\{\tau_j\})$ that it judges the Stage~1 skill to have measurably underexplained, and a second pool of workers $W'_i$ is dispatched against $R'$ under the same protocol as (a). The targets are scoped to UI \emph{regions} rather than to specific tasks, so the patch transfers to any test task that touches the same UI surface. Appendix~\ref{app:examples-stage2b} walks through one targeted example.

The combined note set $N = \{n_i\}_{i=1}^{K} \cup \{n'_i\}_{i \,:\, r'_i \in R'}$ is consumed by an assembler agent that consolidates each region's notes into a per-region reference section, an LLM mapper that decides which existing topic $t \in \mathcal{T}$ each region most directly enriches, and an inline step that appends the consolidated reference into the corresponding per-topic guides of both skills via the same joint-generation call. The Stage~2 pair is then $\mathcal{S}^{\mathrm{mm}}_2 = \mathrm{Assemble}(\mathcal{S}^{\mathrm{mm}}_1, N)$ and $\mathcal{S}^{\mathrm{txt}}_2 = \mathrm{Assemble}(\mathcal{S}^{\mathrm{txt}}_1, N)$. The Stage~2 skill shares the topic set $\mathcal{T}$ of its Stage~1 version by construction; only the per-topic guides are augmented. $\mathcal{S}^{\mathrm{mm}}_2$ gains agent-captured screenshots in its $F_t$ and an extended $p_t^{\mathrm{mm}}$ written alongside them. The control $\mathcal{S}^{\mathrm{txt}}_2$ likewise gains an extended $p_t^{\mathrm{txt}}$ that absorbs the same new UI information into words.

\section{Experiments}
\label{sec:experiments}

\subsection{Experimental Setup}
\label{sec:exp-setup}

\paragraph{Agent.}
We evaluate a computer-use agent built on the Claude Code CLI and backed by Claude Opus~4.6. It observes the desktop through screenshots, acts via a fixed GUI tool set, and consults skills through the \texttt{load\_topic} MCP tool from Section~\ref{sec:method-loading}. The CLI, the tool surface, and the per-step prompt are held fixed across all conditions, so any difference in task success is attributable to the skill artefact. We use each benchmark's own per-task budget: CUA-World ships a per-task step cap in \texttt{init.max\_steps} that ranges from $40$ to $200$ actions depending on task difficulty, which we use unchanged; OSExpert-Eval is wall-clock-budgeted, and we run it with the benchmark's default $15$~minute per-task cap. The full agent setup --- the GUI tool signatures and the sandboxing setup --- is deferred to Appendix~\ref{app:agent-details}.

\paragraph{Benchmarks.}
We evaluate on two computer-use benchmarks: \textbf{CUA-World}~\citep{aggarwal2026gym} (five domains we use --- LibreOffice Writer / Calc / Impress, QGIS, OpenToonz --- with per-application train/test splits used for both Stage~2 patching and held-out evaluation) and \textbf{OSExpert-Eval}~\citep{liu2026osexpert} (three additional domains --- LibreOffice, GIMP, Tableau --- covering long-horizon compositional workflows on unseen UIs). We use the verifier shipped with each benchmark unchanged. Per-application task counts and train/test splits are listed in Appendix~\ref{app:dataset-details}.

\paragraph{Metric.}
We report the per-task score averaged within each domain, with scores normalised to $[0,1]$ so the two benchmarks are directly comparable. CUA-World scores each task with a checklist-based VLM verifier that decomposes the task into weighted subtasks and assigns partial credit~\citep{aggarwal2026gym}: we take the average checklist score per task, rescaled from $0$--$100$ to $[0,1]$. OSExpert-Eval scores each task with the deterministic state-based verifier shipped with the benchmark~\citep{liu2026osexpert}, returning $1$ on success and $0$ otherwise; the per-domain mean therefore reduces to the domain success rate.

\paragraph{Skill conditions.}
We compare five conditions:
\begin{itemize}[leftmargin=*,topsep=2pt,itemsep=2pt]
\item \textbf{No-skill}: agent runs with only the task instruction, with no skill provided.
\item \textbf{Stage~1 \ours{}}, with the \textbf{Stage~1 text-only control} for comparison: both mined from authored documentation.
\item \textbf{Stage~2 \ours{}}, with the \textbf{Stage~2 text-only control} for comparison: both after Stage~2 enrichment via UI exploration.
\end{itemize}
Within each stage, the multimodal skill and its text-only control are generated jointly from the same source content (Section~\ref{sec:method-construct}), so any performance gap between them isolates the modality's effect.

\subsection{Main Results}
\label{sec:exp-main}

\definecolor{banner}{HTML}{DCE9F4}   
\definecolor{altrow}{HTML}{F2F2F2}   

\begin{table*}[t]
\centering
\scriptsize
\setlength{\tabcolsep}{4pt}
\renewcommand{\arraystretch}{1.25}
\begin{tabular}{l ccc c c ccccc c c c}
\toprule
   & \multicolumn{4}{c}{\textbf{OSExpert-Eval}}
   &
   & \multicolumn{6}{c}{\textbf{CUA-World}}
   &
   & \textbf{All} \\
\cmidrule(lr){2-5} \cmidrule(lr){7-12}
\textbf{Method}
   & LibreOffice & GIMP & Tableau & \textbf{Avg}
   &
   & Writer & Calc & Impress & QGIS & OpenToonz & \textbf{Avg}
   &
   & \textbf{Avg} \\
   & {\scriptsize (24)} & {\scriptsize (6)} & {\scriptsize (20)} &
   &
   & {\scriptsize (24)} & {\scriptsize (38)} & {\scriptsize (28)} & {\scriptsize (16)} & {\scriptsize (21)} &
   &
   & \\
\midrule
No-skill baseline
   & 0.292 & 0.167 & 0.100 & 0.186
   &
   & 0.135 & 0.512 & 0.345 & 0.660 & 0.210 & 0.372
   &
   & 0.303 \\
\midrule

\rowcolor{banner}
\multicolumn{14}{l}{\hspace*{-2pt}\textit{\textbf{Stage 1}} (doc-derived)} \\

\hspace*{1em}Text
   & 0.333 & 0.167 & 0.150 & 0.217
   &
   & 0.153 & 0.603 & 0.505 & 0.715 & 0.124 & 0.420
   &
   & 0.344 \\

\rowcolor{altrow}
\hspace*{1em}\ours
   & 0.375 & 0.167 & 0.200 & 0.247
   &
   & 0.228 & 0.585 & 0.496 & 0.706 & 0.145 & 0.432
   &
   & 0.363 \\
\midrule

\rowcolor{banner}
\multicolumn{14}{l}{\hspace*{-2pt}\textit{\textbf{Stage 2}} (+ UI explorer)} \\

\hspace*{1em}Text
   & 0.417 & 0.167 & 0.150 & 0.245
   &
   & 0.225 & 0.620 & 0.509 & 0.708 & 0.185 & 0.449
   &
   & 0.373 \\

\rowcolor{altrow}
\hspace*{1em}\ours
   & \textbf{0.500} & \textbf{0.333} & \textbf{0.300} & \textbf{0.378}
   &
   & \textbf{0.276} & \textbf{0.656} & \textbf{0.580} & \textbf{0.726} & \textbf{0.274} & \textbf{0.502}
   &
   & \textbf{0.456} \\
\bottomrule
\end{tabular}
\caption{Main results across the two computer-use benchmarks our method targets. Each cell is the per-domain mean score in $[0,1]$; the integer beneath each domain header is the task count. \textbf{Stage~1} skills are mined from documentation; \textbf{Stage~2} skills additionally incorporate the UI explorer output. For each stage we report two matched variants: multimodal (figures retained) and a text-only control. Per-benchmark \textbf{Avg} columns are unweighted means over the domains within each benchmark; the \textbf{All Avg} column is the unweighted mean over all domains.}
\label{tab:main}
\end{table*}

\paragraph{Skill helps; UI exploration helps more.}
Adding any skill over the no-skill baseline improves performance on every domain we measure. Averaged unweighted across the eight domains in Table~\ref{tab:main}, the no-skill agent scores $0.303$. Stage~1 \ours lifts it to $0.363$ ($+6.0\%$ absolute), and Stage~2 \ours --- adding UI exploration on top of Stage~1 --- lifts it further to $0.456$ ($+9.3\%$ over Stage~1, $+15.3\%$ over no-skill). The Stage~2 gain concentrates on domains whose UI is most under-documented by the manual: GIMP ($+16.6\%$ absolute over Stage~1 \ours{}) and OpenToonz ($+12.9\%$). On domains whose documentation already covers most controls (e.g.\ QGIS, Tableau), Stage~2 closes a smaller gap, consistent with the design intent of UI exploration as a complement to authored knowledge rather than a replacement.

\paragraph{\ours{} outperforms its matched text-only control, especially after UI exploration.}
Within each stage, the multimodal variant beats its matched text-only control on $5/8$ domains at Stage~1 (and ties on GIMP) and on $8/8$ at Stage~2. Averaged across the eight domains, the multimodal gain is modest at Stage~1 ($+1.9\%$ absolute, $0.344 \to 0.363$) but substantially larger at Stage~2 ($+8.3\%$ absolute, $0.373 \to 0.456$). The largest Stage~2 multimodal lifts appear on the visually intensive creative tools (GIMP $+16.6\%$, OpenToonz $+8.9\%$, OSExpert Tableau $+15.0\%$) and the OSExpert LibreOffice subset ($+8.3\%$), where success depends heavily on recognising dialogs, icons, and UI layouts. The gain shrinks but does not disappear on office-productivity workflows --- CUA-World Writer ($+5.1\%$ at Stage~2), Calc ($+3.6\%$), Impress ($+7.1\%$). Retaining figures helps most where visual grounding is the bottleneck, and gives smaller but consistent gains elsewhere.

\section{Analysis}
\label{sec:analysis}

\subsection{When Do Figures Help?}
\label{sec:exp-qualitative}

Across the cases where \ours{} outperforms its text-only control, the multimodal advantage concentrates on the two failure modes of text-only skills we anticipated in the introduction.

\textbf{(i) Identifying the right UI element to act on.}
We see three recurring situations in which a reference screenshot is hard to substitute with text. The first is graphical controls whose meaning lives in their appearance: toolbar icons, palette swatches, drawing-tool glyphs. A text description such as ``the brush tool'' or ``the format-paint icon'' is easy to confuse with a neighbouring button, since icon-based controls do not carry text labels. The second is targets that only appear after an earlier interaction: modal sub-dialogs two menus deep, expanded dropdowns, right-click context menus, and pop-up panels. These surfaces are not visible in the idle screenshot, and a text guide can only describe them in the abstract. The third is a sub-region of a larger visual element --- a text field next to its drop-arrow, a button in the footer of a tall dialog --- where a phrase like ``click the dropdown'' or ``Save at the bottom of the dialog'' names the surrounding element rather than the actual click target, so the agent often clicks the wrong region. A reference screenshot addresses all three by showing both the target's appearance and its exact location. Appendix~\ref{app:qa-examples} walks through one example in which the text-only agent reads ``Save at the bottom of the dialog'' and clicks the title bar instead, closing the application.

\textbf{(ii) Verifying the UI state between steps.}
Multi-step workflows require the agent to verify, after each action, that the UI has actually reached the expected state before continuing. With a text description, this verification is indirect: the agent has to imagine what the description should look like in pixels, then check whether the current view matches. With a reference screenshot, the comparison is direct. The gap is largest when the failure is silent --- form fields that look like they accepted a typed value but rejected it on commit, sequenced dialogs whose progress is only visible in small layout changes --- because in these cases the agent has no other cue except the screen itself.

\textbf{When figures don't help.} The gap between \ours{} and its text-only control collapses on tasks consisting of short, explicitly specified action sequences, where the required operations can already be communicated clearly through text alone, leaving little additional value for figures to provide.

\subsection{Ablation: MCP Tool vs.\ Plain \texttt{Read}}
\label{sec:exp-loading}

Section~\ref{sec:method-loading} exposes skill loading to the agent as the MCP tool \texttt{load\_topic} rather than letting the agent issue plain \texttt{Read} calls against the skill folder. To test that choice empirically, we ablate the MCP tool against a single \textbf{Direct \texttt{Read}} baseline: the system prompt still asks the agent to consult the skill at every step, but instead of calling \texttt{load\_topic} the agent uses the plain \texttt{Read} tool, reading \texttt{guide.md} for the relevant topic and then issuing one further \texttt{Read} for each figure it judges worth fetching. Both methods use the same agent setup and the same skill artefact; only how the agent retrieves content from the skill folder differs.


\providecolor{altrow}{HTML}{F2F2F2}

\begin{table*}[t]
\centering
\small
\setlength{\tabcolsep}{8pt}
\renewcommand{\arraystretch}{1.25}
\begin{tabular}{l ccc c ccc}
\toprule
  &                    &                          &                      & & \multicolumn{3}{c}{\textbf{CUA-World \ours{} test score}} \\
\cmidrule(lr){6-8}
\textbf{Method} & \textbf{Load rate} & \textbf{Figures / task} & \textbf{Last @ step} & & Writer & OpenToonz & QGIS \\
                &                    &                          &                      & & {\scriptsize (24)} & {\scriptsize (21)} & {\scriptsize (16)} \\
\midrule
Direct \texttt{Read}        & $92.6\%$         & $0.8$         & $1.5$            & & $0.236$ & $0.246$ & 0.695 \\
\rowcolor{altrow}
\textbf{MCP tool (ours)}    & $\mathbf{100\%}$ & $\mathbf{7.9}$ & $\mathbf{10.4}$ & & $\mathbf{0.276}$ & $\mathbf{0.274}$ & 0.726 \\
\bottomrule
\end{tabular}
\caption{Skill-loading ablation. \emph{Load rate}, \emph{Figures / task}, and \emph{Last @ step} are measured on the Writer training split with the Stage~2 multimodal skill : \emph{Load rate} is the fraction of trajectories that consult the skill at all; \emph{Figures / task} is the average number of figures delivered per task; \emph{Last @ step} is the median trajectory step of the final skill consultation. \emph{CUA-World \ours{} test score} is the Stage~2 multimodal mean on each domain; the integer beneath each header is the task count. Both methods use the same agent setup and the same skill artefact and differ only in how the agent retrieves content from the skill folder.}
\label{tab:loading-diag}
\end{table*}

\paragraph{What the agent does differently.}
On the Writer training split, Direct \texttt{Read} reaches the skill on $92.6\%$ of tasks but reads only $0.8$ figures per task on average, and the median last consult lands at step $1.5$ --- the agent reads the skill once near the start of the rollout and never returns. The MCP tool runs at $100\%$ load rate with $7.9$ figures per task ($\sim$$\mathbf{10\times}$ more) and a median last consult step of $10.4$, so the agent keeps consulting the skill as the task moves between UI surfaces. The gap is structural: each figure under Direct \texttt{Read} costs the agent an extra \texttt{Read} call and is usually skipped, while the MCP tool delivers every referenced figure with the prose in one tool result (Table~\ref{tab:loading-diag}, left).

\paragraph{Effect on accuracy.}
The MCP tool gives a consistent $+\mathbf{3{-}4\%}$ absolute lift over Direct \texttt{Read} across all three domains: Writer ($0.236 \to 0.276$, $+4.0\%$), OpenToonz ($0.246 \to 0.274$, $+2.8\%$), and QGIS ($0.695 \to 0.726$, $+3.1\%$). The multimodal skill carries its UI knowledge in the figures, and without the MCP tool most of that knowledge is left on the table.

\subsection{Contribution of Each Phase of \ours}
\label{sec:exp-phase-ablation}

\ours{} is built in two stages (Section~\ref{sec:method-construct}): \textbf{Stage~1} mines the application's authored documentation, and \textbf{Stage~2} enriches the resulting skill with screenshots and notes captured by driving the live application, in two sub-passes --- \textbf{S2(a)}, a free UI explorer that visits every region of the idle window, and \textbf{S2(b)}, a trajectory-targeted explorer that revisits UI regions where training-task rollouts failed. Table~\ref{tab:phase-ablation} reports each phase's contribution.


\providecolor{banner}{HTML}{DCE9F4}
\providecolor{altrow}{HTML}{F2F2F2}

\begin{table}[h]
\centering
\scriptsize
\setlength{\tabcolsep}{5pt}
\renewcommand{\arraystretch}{1.25}
\begin{tabular}{l cccc}
\toprule
\textbf{Method}
  & Writer & Calc & Impress
  & \textbf{Avg} \\
  & {\scriptsize (24)} & {\scriptsize (38)} & {\scriptsize (28)}
  & \\
\midrule
No-skill baseline             & 0.135 & 0.512 & 0.345 & 0.331 \\
\midrule

\rowcolor{banner}
\multicolumn{5}{l}{\hspace*{-2pt}\textit{\textbf{Stage 1}} (doc-derived)} \\
\hspace*{1em}Text             & 0.153 & 0.603 & 0.505 & 0.420 \\
\rowcolor{altrow}
\hspace*{1em}\ours            & 0.228 & 0.585 & 0.496 & 0.436 \\
\midrule

\rowcolor{banner}
\multicolumn{5}{l}{\hspace*{-2pt}\textit{\textbf{Stage 2}} (+ UI explorer)} \\
\hspace*{1em}+ S2(a) only (\ours) & 0.236 & 0.608 & 0.543 & 0.462 \\
\hspace*{1em}+ S2(b) only (\ours) & 0.255 & 0.600 & 0.505 & 0.453 \\
\hspace*{1em}Text (full)          & 0.225 & 0.620 & 0.509 & 0.451 \\
\rowcolor{altrow}
\hspace*{1em}\ours (full)         & \textbf{0.276} & \textbf{0.656} & \textbf{0.580} & \textbf{0.504} \\
\bottomrule
\end{tabular}
\caption{Phase ablation on CUA-World LibreOffice. \emph{Stage 1} variants are mined from authored documentation. \emph{Stage 2} variants additionally incorporate the UI explorer output: \textbf{S2(a)} = free explorer only; \textbf{S2(b)} = trajectory-targeted explorer only; \emph{full} = both sub-passes. Integers below domain names give task counts.}
\label{tab:phase-ablation}
\end{table}

\paragraph{Stage 1 is the backbone.}
With Stage~1 alone, the LibreOffice average jumps from $0.331$ (no-skill) to $0.436$ (\ours) --- a $\mathbf{+10.5\%}$ absolute gain, confirming that mature applications' authored documentation already covers a large fraction of the procedural knowledge an agent needs. Within Stage~1, however, the multimodal-over-text gap is small ($+1.6\%$ absolute, $0.420 \to 0.436$): the manual's content is mostly procedural --- menu paths, dialog field names, keyboard shortcuts --- which text describes clearly on its own; the figures it ships are vendor renders that add little to the text.

\paragraph{Stage 2(a) and 2(b) patch complementary surfaces.}
Each Stage~2 sub-pass adds $\sim$$\mathbf{2{-}3\%}$ absolute on top of Stage~1 ($0.436 \to 0.462$ for 2(a) alone, $0.436 \to 0.453$ for 2(b) alone), but they patch \emph{different} parts of the UI. Stage~2(a) (free explorer) reaches the \emph{static} surfaces visible in the idle window --- toolbars, sidebar decks, menus, palette decks; the planner partitions what is on-screen at rest and dispatches a worker per region. Stage~2(b) (trajectory-targeted) instead targets \emph{dynamic} surfaces that only appear during real task interactions --- modal dialogs reached two menus deep, pop-up state-transition behaviours (e.g.\ the margin spinbox rejecting unit suffixes, Section~\ref{sec:exp-qualitative}), and post-action confirmation frames. Combining both is super-additive: the full Stage~2 \ours{} reaches $0.504$, beating either sub-pass by $\mathbf{+4{-}5\%}$ absolute and the matched text-only control by $+5.3\%$ ($0.451 \to 0.504$). We adopt the full Stage~2 multimodal configuration as the default reported in Table~\ref{tab:main}.

\section{Related Work}
\label{sec:related_work}


\paragraph{Computer-Use Agents.}
Computer-use agents (CUAs) operate directly on graphical user interfaces to complete user-specified tasks~\citep{agashe2025agent, xie2024osworld, wang2026opencua}.
Unlike API-based agents that interact with structured tool surfaces, CUAs must perceive the screen through screenshots and act through low-level mouse and keyboard primitives, which gives them broad applicability across software but makes every step dependent on visual grounding.
Recent systems have approached human-level performance on standardised benchmarks~\citep{claude_sonnet_46, agent_s3}, yet they generalise poorly to unseen UIs and long-horizon workflows~\citep{aggarwal2026gym, liu2026osexpert}, because the application-specific procedural knowledge required for such tasks --- which menu hides a given command, what dialog appears after a click, how a multi-step widget transitions between states --- is not directly supplied by pre-training. This leaves room for external, application-specific knowledge sources to be consulted at decision time.

\paragraph{Skill Libraries for CUA.}
Recent work equips agents with reusable skill libraries to supply application-specific procedural knowledge at decision time~\citep{chen2026cua, liu2026osexpert, jiang2026xskill, wang2025inducing}.
CUA-Skill~\citep{chen2026cua} encodes computer-use knowledge as parameterised execution and composition graphs spanning common Windows applications. OSExpert~\citep{liu2026osexpert} learns unit skills through GUI-based depth-first search exploration of the live environment and composes them into curricula for complex tasks. XSkill~\citep{jiang2026xskill} treats skills as cross-task transferable behaviours learned through imitation, while~\citet{wang2025inducing} induce structured skills directly from agent rollouts on solved tasks.
Across all of these, however, the skill artifact itself is consistently text: natural-language intents, action specifications, or symbolic procedures, with no first-class slot for figures or screenshots.
This is a poor fit for GUI environments: many UI elements are ambiguous to describe in text, and agents cannot directly match a textual state description against their current screenshot observation. Verifying intermediate state in multi-step workflows is similarly hampered when the expected post-action UI can only be described in words rather than shown as a reference frame.

\paragraph{Concurrent Work.}
A recently published work, MMSkills~\citep{zhang2026mmskills}, represents each skill as a per-subtask package coupling a textual procedure with runtime state cards and multi-view screenshots extracted from public trajectories, and introduces branch loading to align skill evidence with the live environment in a temporary inference branch.
\ours{} differs from MMSkills in skill granularity, construction source, and loading mechanism: it scopes one skill per application organised as a hierarchical topic tree, builds skills from authored vendor documentation and live UI exploration rather than trajectory mining, and exposes an MCP tool that the agent can call to fetch relevant topics on demand rather than performing branch-based environment alignment. These design choices are deliberate: per-application hierarchical organisation lets a single skill scale to the full surface of a complex application, authored documentation provides a coverage backbone that trajectory mining alone cannot match, and an MCP tool gives the agent uniform access to text and figures within its existing tool-calling interface, without spawning a separate inference branch.
\section{Conclusion}
\label{sec:conclusion}

We introduced \ours, a hierarchical multimodal skill artifact for computer-use agents that retains UI figures alongside text and exposes them through an on-demand \texttt{load\_topic} MCP tool. \ours{} skills are built by a two-stage pipeline --- documentation mining followed by complementary free and trajectory-targeted UI exploration. To measure the contribution of the visual modality, each \ours{} skill is paired with a text-only control generated jointly from the same source content. Across $177$ tasks from CUA-World and OSExpert-Eval, \ours{} lifts a Claude Opus~4.6 agent from a no-skill mean of $0.303$ to $0.456$ ($+15.3\%$ absolute), outperforming its text-only control by $+8.3\%$ absolute at Stage~2. A qualitative audit pinpoints three categories of UI knowledge for which figures are particularly load-bearing --- where to click, whether a typed value committed, and how to operate a multi-step widget --- and ablations show that both the MCP-based delivery mechanism and the trajectory-targeted UI exploration are individually necessary to the headline gain. We see \ours{} as evidence that the skill artifact for computer-use agents should be a multimodal, on-demand reference rather than a text-only document, and we hope the open-source pipeline and controlled evaluation protocol make further work on figure-grounded skills straightforward.

\section*{Limitations}
We observe that current models do not reliably decide which figures within a loaded topic are relevant to their immediate decision, and underutilize the figures delivered by multimodal skills. Given this, our \texttt{load\_topic} mechanism returns the full set of figures associated with a topic in a single tool result. However, this coarse granularity inflates the per-call context cost and, on weaker base models, can outweigh the multimodal benefit: as reported in Appendix~\ref{app:qwen3}, \textsc{Qwen3.5-397B-A17B-FP8} on OSWorld LibreOffice degrades to no-skill levels under the multimodal condition, with trajectories exhibiting malformed tool calls consistent with context-length pressure. How to deliver multimodal skill content within a more effective context budget remains an open question.



\bibliography{custom}

\appendix

\clearpage

\lstdefinestyle{promptcode}{
  basicstyle=\ttfamily\footnotesize,
  breaklines=true,
  breakatwhitespace=false,
  breakindent=0pt,
  columns=fullflexible,
  keepspaces=true,
  showstringspaces=false,
  upquote=true,
  frame=none,
  xleftmargin=0pt,
  xrightmargin=0pt,
  aboveskip=2pt,
  belowskip=2pt,
}

\newtcolorbox{promptbox}[2][]{
  enhanced, breakable, sharp corners,
  colback=gray!4, colframe=gray!70,
  fonttitle=\bfseries\footnotesize,
  coltitle=white, colbacktitle=black!75,
  title={#2}, attach boxed title to top left={xshift=0pt, yshift*=-1mm},
  boxed title style={sharp corners, colback=black!75, colframe=black!75},
  left=2mm, right=2mm, top=3.5mm, bottom=2mm,
  #1
}

\section*{Appendix Contents}
{\small\setlength{\parindent}{0pt}\setlength{\parskip}{1pt}

\textbf{A}\hspace{0.6em}\textbf{Skill Overview}\dotfill \pageref{app:skill-overview}\\
\hspace*{1.6em}A.1\hspace{0.6em}Structure\dotfill \pageref{app:skill-overview-structure}\\
\hspace*{1.6em}A.2\hspace{0.6em}\ours{} and Text-Only Control: Side-by-Side\dotfill \pageref{app:skill-overview-matched}

\vspace{3pt}
\textbf{B}\hspace{0.6em}\textbf{Inference-Time Skill Loading}\dotfill \pageref{app:inference-loading}

\vspace{3pt}
\textbf{C}\hspace{0.6em}\textbf{Three Construction-Stage Examples}\dotfill \pageref{app:examples}\\
\hspace*{1.6em}C.1\hspace{0.6em}Stage 1: From Authored Documentation\dotfill \pageref{app:examples-stage1}\\
\hspace*{1.6em}C.2\hspace{0.6em}Stage 2(a): Free UI Explorer\dotfill \pageref{app:examples-stage2a}\\
\hspace*{1.6em}C.3\hspace{0.6em}Stage 2(b): Trajectory-Targeted Explorer\dotfill \pageref{app:examples-stage2b}

\vspace{3pt}
\textbf{D}\hspace{0.6em}\textbf{Agent Setup}\dotfill \pageref{app:agent-details}

\vspace{3pt}
\textbf{E}\hspace{0.6em}\textbf{Datasets}\dotfill \pageref{app:dataset-details}

}
\bigskip

%
%

\graphicspath{{sections/appendix/appendix_skill_overview/figures/}{appendix/appendix_skill_overview/figures/}{appendix_skill_overview/figures/}{figures/}{./}}

\section{Skill Overview}
\label{app:skill-overview}

This appendix grounds the formal notation of Section~\ref{sec:method-def} (the topic set $\mathcal{T}$, the per-topic guide $g_t = (p_t, F_t)$, the central \texttt{SKILL.md} index, the \ours{} skill $\mathcal{S}^{\mathrm{mm}}$, and its text-only control $\mathcal{S}^{\mathrm{txt}}$) in a real skill (the Stage~2 Writer skill we ship with the paper). Section~\ref{app:skill-overview-structure} shows the file tree and explains how each notation element maps to a file or folder. Section~\ref{app:skill-overview-matched} shows the \ours{} skill and its text-only control on one concrete topic side by side.

\subsection{Structure}
\label{app:skill-overview-structure}

A skill is laid out as a centralised \texttt{SKILL.md} index plus a folder per topic, arranged in the two-layer hierarchy of Section~\ref{sec:method-def}: top-level category folders group related topics, and each leaf topic folder holds the guide. Figure~\ref{lst:skill-tree} shows the layout for our Writer skill (two categories are fully expanded for illustration; there are 19 categories in total).

\begin{figure}[h]
\centering
\begin{tcolorbox}[colback=gray!4, colframe=gray!50, sharp corners,
                  boxrule=0.5pt, left=6pt, right=6pt, top=4pt, bottom=4pt,
                  fontupper=\ttfamily\scriptsize]
.\\
|-{}- SKILL.md\\
|-{}- introducing-writer/\\
|~~~|-{}- writer-interface/\\
|~~~|~~~|-{}- guide.md\\
|~~~|~~~`-{}- fig01.png, fig02.png\\
|~~~|-{}- creating-and-opening-documents/\\
|~~~|~~~`-{}- guide.md\\
|~~~`-{}- \ldots\\
|-{}- formatting-text/\\
|~~~|-{}- character-formatting/\\
|~~~|~~~|-{}- guide.md\\
|~~~|~~~`-{}- fig01.png, fig02.png, ui-character-dialog.png, \ldots\\
|~~~|-{}- paragraph-formatting/\\
|~~~`-{}- \ldots\\
|-{}- page-layout-basics/\\
|-{}- introduction-to-styles/\\
|-{}- working-with-styles/\\
`-{}- \ldots
\end{tcolorbox}
\caption{File layout of the Writer skill (root folder omitted; two categories fully expanded for illustration, $19$ categories total). \texttt{SKILL.md} is the central index. Each leaf folder is a topic $t \in \mathcal{T}$ and carries its $g_t = (p_t, F_t)$ as a \texttt{guide.md} (the text body $p_t$) plus a set of PNG files (the figures $F_t$). For instance, the topic \texttt{formatting-text/character-formatting} has $p_t = $ its \texttt{guide.md} and $F_t = \{$\texttt{fig01.png}, \texttt{fig02.png}, \texttt{ui-character-dialog.png}, \ldots$\}$.}
\label{lst:skill-tree}
\end{figure}

\begin{figure}[h]
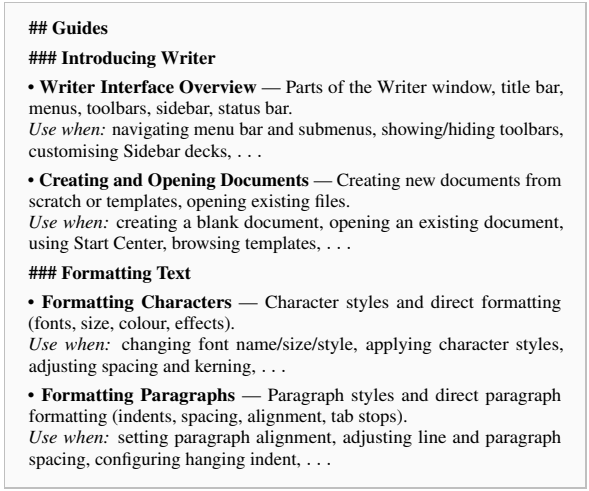

\centering
\begin{tcolorbox}[colback=gray!4, colframe=gray!50, sharp corners,
                  boxrule=0.5pt, left=6pt, right=6pt, top=4pt, bottom=4pt,
                  fontupper=\scriptsize]
\textbf{\#\#\ Guides}

\smallskip
\textbf{\#\#\#\ Introducing Writer}

\smallskip
\textbullet\ \textbf{Writer Interface Overview} --- Parts of the Writer window, title bar, menus, toolbars, sidebar, status bar.\\
\textit{Use when:} navigating menu bar and submenus, showing/hiding toolbars, customising Sidebar decks, $\ldots$

\smallskip
\textbullet\ \textbf{Creating and Opening Documents} --- Creating new documents from scratch or templates, opening existing files.\\
\textit{Use when:} creating a blank document, opening an existing document, using Start Center, browsing templates, $\ldots$

\smallskip
\textbf{\#\#\#\ Formatting Text}

\smallskip
\textbullet\ \textbf{Formatting Characters} --- Character styles and direct formatting (fonts, size, colour, effects).\\
\textit{Use when:} changing font name/size/style, applying character styles, adjusting spacing and kerning, $\ldots$

\smallskip
\textbullet\ \textbf{Formatting Paragraphs} --- Paragraph styles and direct paragraph formatting (indents, spacing, alignment, tab stops).\\
\textit{Use when:} setting paragraph alignment, adjusting line and paragraph spacing, configuring hanging indent, $\ldots$
\end{tcolorbox}
\caption{Representative excerpt of the central \texttt{SKILL.md} (verbatim from the Stage~2 Writer skill; markdown link syntax simplified to plain titles). Each item names one topic and gives the \textit{when to use} criterion that the agent matches against its current task before invoking \texttt{load\_topic} (Section~\ref{sec:method-loading}). The agent reads only this index up front; per-topic content is fetched on demand.}
\label{lst:skill-index}
\end{figure}

\paragraph{Mapping the notation.}
Each element of Section~\ref{sec:method-def}'s definition has a direct realisation in the file layout:
\begin{itemize}[leftmargin=*,topsep=2pt,itemsep=0pt]
  \item $\mathcal{T}$ is the set of leaf topic folders. A topic identifier is its path slug, e.g.\ \texttt{formatting-text/character-formatting}.
  \item $g_t = (p_t, F_t)$ is the contents of a leaf folder: \texttt{guide.md} is the text body $p_t$, and every PNG file in the same folder is an element of $F_t$.
  \item The central \texttt{SKILL.md} is the index that the agent reads first. Each entry points to one per-topic guide and tags it with a short \emph{when to use} description; the agent matches its current task against these descriptions to decide which $g_t$ to load via \texttt{load\_topic} (Section~\ref{sec:method-loading}). Figure~\ref{lst:skill-index} reproduces a representative excerpt.
\end{itemize}

\subsection{\ours{} and Its Text-Only Control: One Topic in Two Modalities}
\label{app:skill-overview-matched}

\begin{figure*}[!t]
  \centering
  \begin{subfigure}[t]{0.49\textwidth}
    \begin{tcolorbox}[colback=orange!5, colframe=orange!50!black, sharp corners,
                      boxrule=0.6pt, left=5pt, right=5pt, top=4pt, bottom=4pt,
                      title={\scriptsize\textbf{$\mathcal{S}^{\mathrm{txt}}$ \textbullet\ \texttt{character-formatting/guide.md}}},
                      coltitle=white, colbacktitle=orange!50!black,
                      fonttitle=\bfseries,
                      equal height group=skov-mm-vs-txt]
\scriptsize
\textbf{\# Formatting Characters}\\[3pt]
Direct formatting works well for one-off changes. With text selected, the controls live in two places: the Formatting toolbar at the top of the window, and the Character section of the Sidebar's Properties deck. Both surfaces expose the same core attributes --- font name, font size, bold, italic, underline, strikethrough, font and highlight colours, and super/subscript.\\[3pt]
\textit{Layout of the Character section in the Properties deck, top to bottom:} the Font Name dropdown (e.g.\ ``Liberation Sans'') with the Font Size dropdown (``12\,pt'') to its right; a first row of toggle buttons --- Bold, Italic, Underline, Strikethrough, Toggle Shadow, Increase Font Size, Decrease Font Size; a second row --- Font Color, Highlight Color, Clear Direct Formatting, Set Character Spacing, Superscript, Subscript. A \textbf{More Options} link in the top-right of the section opens the full Character dialog.\\[3pt]
For step-changes in size, the \textbf{Increase Font Size} / \textbf{Decrease Font Size} buttons jump in fixed 2\,pt increments; for an exact value, type into the Font Size dropdown directly.
\end{tcolorbox}
  \end{subfigure}\hfill
  \begin{subfigure}[t]{0.49\textwidth}
    \begin{tcolorbox}[colback=cyan!4, colframe=cyan!50!black, sharp corners,
                      boxrule=0.6pt, left=5pt, right=5pt, top=4pt, bottom=4pt,
                      title={\scriptsize\textbf{$\mathcal{S}^{\mathrm{mm}}$ \textbullet\ \texttt{character-formatting/guide.md}}},
                      coltitle=white, colbacktitle=cyan!50!black,
                      fonttitle=\bfseries,
                      equal height group=skov-mm-vs-txt]
\scriptsize
\textbf{\# Formatting Characters}\\[3pt]
For one-off tweaks, direct formatting is the fastest path. Select your text and reach for either the Formatting toolbar or the Character section of the Sidebar's Properties deck --- font name, size, bold/italic/underline/strikethrough, font and highlight colours, and super/subscript are all there, plus a \textbf{More Options} link that opens the full Character dialog.\\[3pt]
\textit{See} \texttt{`fig01.png`}\textit{ for the Character section layout:}\\[2pt]
\begin{center}
\includegraphics[width=0.95\linewidth]{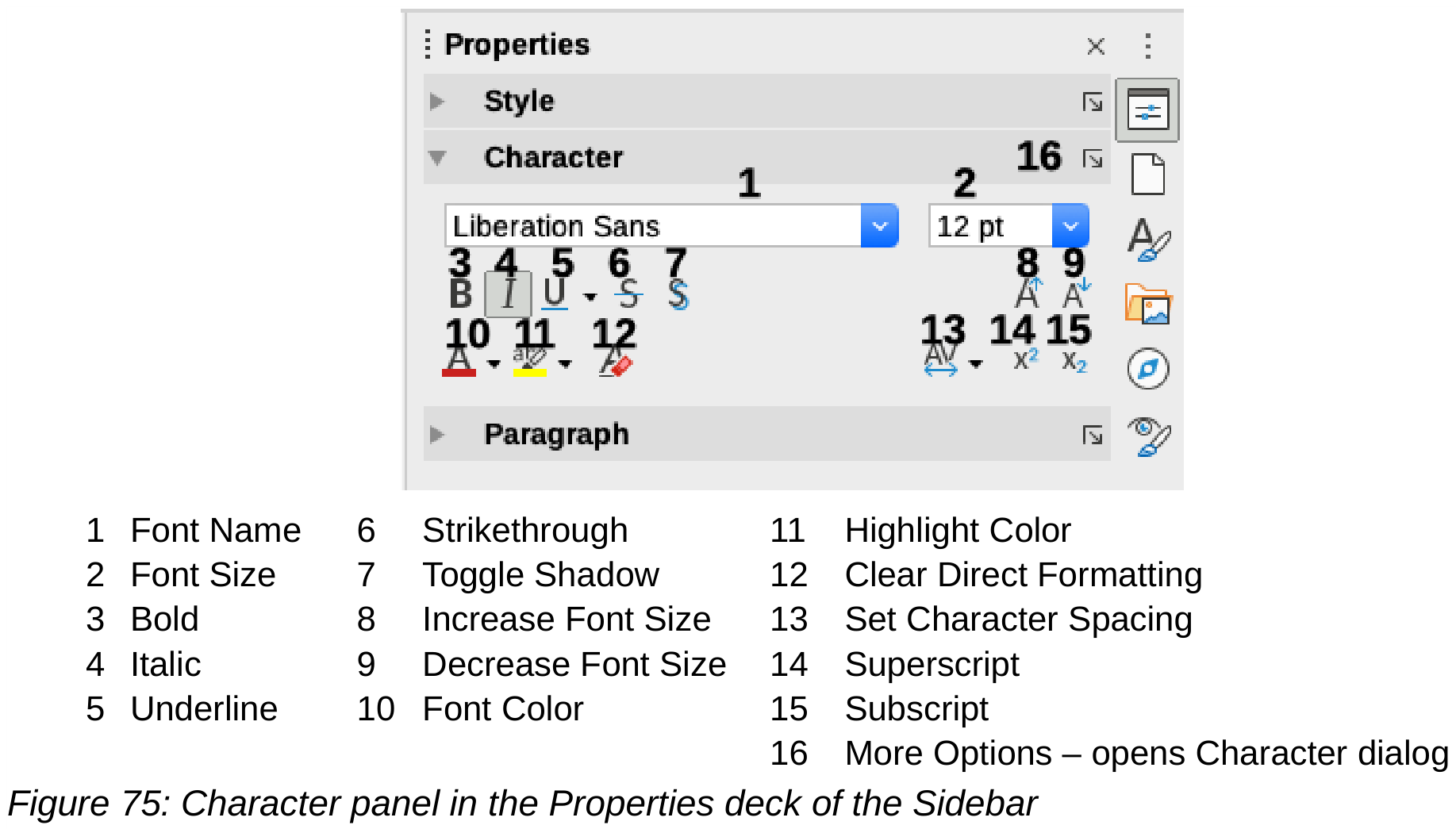}
\end{center}
~\\[-4pt]
To step font size up or down, use the \textbf{Increase Font Size} / \textbf{Decrease Font Size} buttons (fixed 2\,pt increments); for precise sizing, type into the Font Size dropdown.
\end{tcolorbox}
  \end{subfigure}
  \caption{\ours{} and its text-only control on topic $t = $ \texttt{formatting-text/character-formatting}. Both panels are real \texttt{guide.md} bodies and were produced in the same LLM call from the same source context (Section~\ref{sec:method-construct}); they cover the same procedural content but their text is not word-for-word identical. \textbf{Left} ($\mathcal{S}^{\mathrm{txt}}$): the text-only control carries no figure ($F_t^{\mathrm{txt}} = \emptyset$); the second paragraph absorbs the layout of the Character section into words. \textbf{Right} ($\mathcal{S}^{\mathrm{mm}}$): the multimodal skill references \texttt{fig01.png} inline and the figure is an element of $F_t^{\mathrm{mm}}$, so its text can defer that layout to the figure and stays shorter. The figure presence and the wording of $p_t$ around it are the only systematic differences between the two variants.}
  \label{fig:skov-mm-vs-txt}
\end{figure*}

To make the control construction of Section~\ref{sec:method-def} concrete, we show the \ours{} skill $\mathcal{S}^{\mathrm{mm}}$ and its text-only control $\mathcal{S}^{\mathrm{txt}}$ on one specific topic, $t = $ \texttt{formatting-text/character-formatting}. Both are produced in the \emph{same} LLM call from the same input context (Section~\ref{sec:method-construct}), so they cover the same procedural content of $p_t$; the two text bodies are not word-for-word identical, but each is the natural expression of that content within its own modality. Where $\mathcal{S}^{\mathrm{mm}}$ embeds a figure (an element of $F_t$) and references it inline, $\mathcal{S}^{\mathrm{txt}}$ absorbs the same visual information into words and carries no figure ($F_t^{\mathrm{txt}} = \emptyset$). Figure~\ref{fig:skov-mm-vs-txt} shows the two variants side by side.

\paragraph{What is shared, and what differs.}
\begin{itemize}[leftmargin=*,topsep=2pt,itemsep=0pt]
  \item \textbf{Shared between the two:} the \texttt{SKILL.md} index, the topic set $\mathcal{T}$, the topic identifier $t = $ \texttt{formatting-text/character-formatting}, its \textit{when to use} index entry, and the procedural content of $p_t$ (which UI controls are mentioned, in which order, and what they do).
  \item \textbf{Differs:} the set $F_t$ of UI figures (the \ours{} skill retains the screenshots; the text-only control has $F_t = \emptyset$), and the wording of $p_t$ in the regions where $\mathcal{S}^{\mathrm{mm}}$ references a figure (there, $\mathcal{S}^{\mathrm{txt}}$ spells the same visual information out in words so that no figure-slot remains in the text).
\end{itemize}

\noindent At evaluation time, any performance gap between $\mathcal{S}^{\mathrm{mm}}$ and $\mathcal{S}^{\mathrm{txt}}$ on the same task is attributable to the modality of presentation rather than to differences in the underlying content (Section~\ref{sec:method-def}). This controlled construction is the contrast used throughout the experiments in Section~\ref{sec:experiments}.

\clearpage
%
%

\graphicspath{{sections/appendix/appendix_inference_loading/figures/}{appendix_inference_loading/figures/}{sections/appendix/appendix_skill_overview/figures/}{figures/}{./}}

\section{Inference-Time Skill Loading}
\label{app:inference-loading}

This appendix makes the loading mechanism of Section~\ref{sec:method-loading} fully concrete in four artefacts: the system prompt that seeds the agent with the \texttt{SKILL.md} index, the \texttt{load\_topic} MCP tool definition together with the server-side interleaving contract, the literal text-image interleaving that arrives back in the tool result, and a worked agent turn end-to-end. All artefacts are verbatim from the shipped Stage~2 Writer skill (\path{skills/libreoffice_writer-knowledge-multimodal-loader-v1/}); only line wrapping and syntax highlighting are cosmetic.

The agent is launched with a system prompt containing a short preface (reproduced below, verbatim from the \texttt{SKILL.md} header) that explains how to consult the skill, followed by the full \texttt{SKILL.md} index in which every topic is paired with its one-sentence \emph{when to use} criterion (Appendix~\ref{app:skill-overview-structure} reproduces a representative excerpt). No per-topic prose and no figures are loaded up front. At every step the agent matches its current sub-goal against the \emph{when to use} lines and, if any matches, issues \texttt{load\_topic(t)} for the chosen topic.

\begin{promptbox}{System prompt preface}
\begin{lstlisting}[style=promptcode]
You are operating LibreOffice Writer 7.3.7.
A skill is registered for you as a set of MCP
tools (already wired):

  - load_topic(topic): returns the chosen
       topic's guide.md AND every figure
       (PNG) in that topic folder as one
       tool response. Use this instead of
       Read for any *.md or figXX.png file
       inside this skill.
  - list_topics(): returns every topic path
       available, one per line.

Each entry in the TOC below has the form
[Title](<topic>/guide.md). The <topic> part
(the path before /guide.md) is what you
pass to load_topic.

Rules:
  1. Before any GUI action where you are
     unsure of the menu path/dialog/icon,
     find the matching topic in the TOC and
     call load_topic first.
  2. You may call load_topic at any step of
     the trajectory, not only at the start.
     If the task moves into a new area
     (e.g. tables), call load_topic again
     for the new area.
  3. Do NOT issue separate Read calls for
     figXX.png files inside this skill --
     they are delivered by load_topic
     automatically.
\end{lstlisting}
\end{promptbox}

\vspace{0.6em}\noindent The \texttt{load\_topic} tool is a stdio MCP tool registered in the agent's tool list, with signature \texttt{\small load\_topic(topic: str)} $\to$ \texttt{\small list[TextContent\,|\,ImageContent]}, where \texttt{topic} is a relative path from the skill root (e.g.\ \texttt{"page-layout-basics/page-numbering"}). The box below reproduces the docstring shown to the model and the server-side rule that produces the interleaved content list (verbatim from \texttt{tools/skill\_server.py}).

\begin{promptbox}{\texttt{load\_topic} docstring and server-side rule}
\textbf{\footnotesize Docstring shown to the model}
\begin{lstlisting}[style=promptcode]
Atomically load a topic's guide.md plus
every adjacent figure image.

Args:
    topic: relative path from the skill
        root, e.g.
        "page-layout-basics/page-numbering".

Returns guide.md prose interleaved with
the referenced figures, so each figure
is delivered immediately after the
sentence that mentions it (figures are
referenced in the guide as `fig01.png`
etc.). Any figures in the folder that
are not referenced in the guide are
appended at the end. Prefer this over
Read for any *.md file inside this
skill -- it removes the need to issue
separate Read calls for the figures.
\end{lstlisting}

\vspace{4pt}\noindent\rule{\linewidth}{0.3pt}\vspace{2pt}

\textbf{\footnotesize Server-side interleaving rule (pseudocode)}
\begin{lstlisting}[style=promptcode]
text   = read(topic/guide.md)
parts  = []
cursor = 0
for m in regex(`figXX.png`).finditer(text):
    parts.append(TextContent(
        text[cursor : m.end()]))
    parts.append(ImageContent(
        load(m.group())))
    cursor = m.end()
parts.append(TextContent(text[cursor:]))
for f in unreferenced_figures(topic):
    parts.append(ImageContent(f))   # fallback
return parts
\end{lstlisting}
\end{promptbox}

\begin{figure*}[!t]
  \centering
  \begin{subfigure}[t]{0.49\textwidth}
    \begin{tcolorbox}[colback=cyan!4, colframe=cyan!50!black, sharp corners,
                      boxrule=0.6pt, left=5pt, right=5pt, top=4pt, bottom=4pt,
                      title={\scriptsize\textbf{$\mathcal{S}^{\mathrm{mm}}$ result: 5 blocks (interleaved)}},
                      coltitle=white, colbacktitle=cyan!50!black,
                      fonttitle=\bfseries,
                      equal height group=apil-result]
\scriptsize
\texttt{\# formatting-text/character-formatting/guide.md}\\[3pt]
Direct formatting is fine for one-off tweaks: select your text and use the buttons on the Formatting toolbar or the Character panel in the Sidebar's Properties deck. You'll find font name, size, bold, italic, underline, strikethrough, font color, highlight color, superscript, subscript, and more.\\
\textit{See} \texttt{`fig01.png`}.
\begin{center}\includegraphics[width=0.9\linewidth]{fig01}\end{center}
\vspace{-2pt}
To bump font size up/down, use the \textbf{Increase/Decrease Font Size} buttons on the Sidebar (fixed 2\,pt steps); for precise sizing, type into the Font Size dropdown directly. For the full set of options, open \textbf{Format $>$ Character\dots}\\
\textit{See} \texttt{`fig02.png`}.
\begin{center}\includegraphics[width=0.9\linewidth]{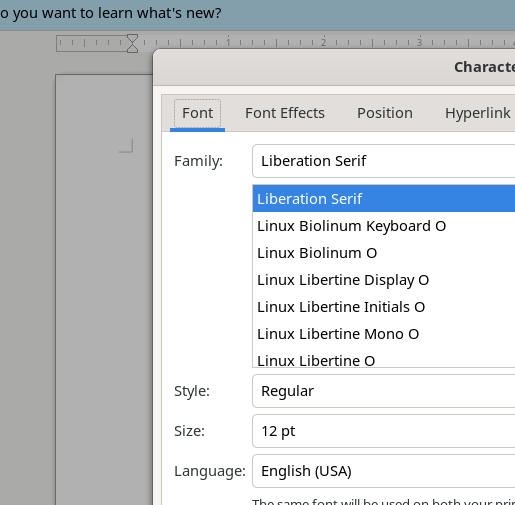}\end{center}
\vspace{-2pt}
Direct character formatting overrides character styles. Use \textbf{Clear Direct Formatting} (Ctrl+M) to strip it out.
\end{tcolorbox}
  \end{subfigure}\hfill
  \begin{subfigure}[t]{0.49\textwidth}
    \begin{tcolorbox}[colback=orange!5, colframe=orange!50!black, sharp corners,
                      boxrule=0.6pt, left=5pt, right=5pt, top=4pt, bottom=4pt,
                      title={\scriptsize\textbf{$\mathcal{S}^{\mathrm{txt}}$ result: 1 block (text)}},
                      coltitle=white, colbacktitle=orange!50!black,
                      fonttitle=\bfseries,
                      equal height group=apil-result]
\scriptsize
\texttt{\# formatting-text/character-formatting/guide.md}\\[3pt]
Direct formatting is fine for one-off tweaks: select your text and use the buttons on the Formatting toolbar or the Character panel in the Sidebar's Properties deck. You'll find font name, size, bold, italic, underline, strikethrough, font color, highlight color, superscript, subscript, and more.\\
\textit{Layout of the Character section, top to bottom:} Font Name dropdown (``Liberation Sans''); Font Size dropdown (``12\,pt''); a row of buttons (Bold, Italic, Underline, Strikethrough, Toggle Shadow, Increase/Decrease Font Size); a second row (Font Color, Highlight Color, Clear Direct Formatting, Set Character Spacing, Superscript, Subscript); a \textbf{More Options} link in the top-right opens the full Character dialog.\\[2pt]
To bump font size up/down, use the \textbf{Increase/Decrease Font Size} buttons on the Sidebar (fixed 2\,pt steps); for precise sizing, type into the Font Size dropdown directly. For the full set of options, open \textbf{Format $>$ Character\dots}\\
\textit{Layout of the Character dialog (Font tab):} six tabs across the top (Font, Font Effects, Position, Hyperlink, Highlighting, Borders); Family dropdown (``Liberation Serif''), Style dropdown (``Regular''), Size dropdown (``12\,pt''), Language dropdown; a live preview pane showing the sample text in the selected font; \textbf{OK}, \textbf{Cancel}, \textbf{Help}, \textbf{Reset} buttons along the bottom.\\[2pt]
Direct character formatting overrides character styles. Use \textbf{Clear Direct Formatting} (Ctrl+M) to strip it out.
\end{tcolorbox}
  \end{subfigure}
  \caption{The literal tool-result structure for one topic, in both variants. \textbf{Left} ($\mathcal{S}^{\mathrm{mm}}$): the content list alternates \texttt{TextContent} and \texttt{ImageContent} blocks; each image is delivered immediately after the sentence that names it (e.g.\ \textit{See} \texttt{`fig01.png`}), giving five blocks total for this topic. \textbf{Right} ($\mathcal{S}^{\mathrm{txt}}$): the content list is a single \texttt{TextContent} block; at the same offsets where the multimodal result emits an image, the text-only guide carries an in-place verbal substitute (italicised \textit{Layout of \dots} paragraphs). Both panels show the same procedural content of $p_t$; the figure presence and the wording around it are the only systematic differences.}
  \label{fig:apil-mm-vs-txt-result}
\end{figure*}

\vspace{0.6em}\noindent The alternative to interleaving --- returning $p_t$ as one text block followed by all of $F_t$ as a flat image array --- would force the model to re-bind each figure to the sentence that introduces it from the trailing image strip. Placing each figure adjacent to its referencing sentence matches how a multimodal model attends to vision/text mixtures and removes the binding step entirely. For the text-only control there is nothing to interleave: the tool returns a single text block whose prose absorbs the visual content of $F_t$ in place. Figure~\ref{fig:apil-mm-vs-txt-result} shows what each variant actually looks like on the wire for one topic.

The agent consumes the returned content list as part of its next assistant turn, exactly like any multimodal context: both text and image blocks count as ordinary input, with no separate ``attachment'' channel. Figure~\ref{fig:apil-transcript} traces one full step end-to-end from a real Stage~2 Writer rollout of \texttt{exam\_paper\_formatting} (Appendix~\ref{app:examples-stage2b}): the agent has just selected a section header that should become a Heading~2, scans the \texttt{SKILL.md} index in its system prompt, matches a topic, calls \texttt{load\_topic}, receives the interleaved tool result, and consumes it on its next step.

\begin{figure*}[!t]
  \centering
  \begin{tcolorbox}[colback=cyan!4, colframe=cyan!50!black, sharp corners,
                    boxrule=0.5pt, left=5pt, right=5pt, top=3pt, bottom=3pt,
                    title={\scriptsize\textbf{Turn $k$ \textbullet\ Agent (assistant message)}},
                    coltitle=white, colbacktitle=cyan!50!black,
                    fonttitle=\bfseries]
\scriptsize\ttfamily
I need to apply the Heading 2 paragraph style. Scanning SKILL.md for a\\
\textit{when to use} line that mentions applying paragraph styles \dots\\[1pt]
\hspace*{2em}- [Applying Styles](introduction-to-styles/applying-styles/guide.md)\\
\hspace*{6em}Use when: applying paragraph styles via toolbar or keyboard\\
\hspace*{6em}shortcuts, applying character styles via Sidebar or context\\
\hspace*{6em}menu, applying page styles via Status bar \dots\\[1pt]
That matches. Loading the topic.
  \end{tcolorbox}

  \begin{tcolorbox}[colback=gray!6, colframe=gray!50, sharp corners,
                    boxrule=0.5pt, left=5pt, right=5pt, top=3pt, bottom=3pt,
                    title={\scriptsize\textbf{Tool call (issued by the agent)}},
                    coltitle=white, colbacktitle=gray!50,
                    fonttitle=\bfseries]
\scriptsize\ttfamily
load\_topic(topic="introduction-to-styles/applying-styles")
  \end{tcolorbox}

  \begin{tcolorbox}[colback=yellow!6, colframe=yellow!60!black, sharp corners,
                    boxrule=0.5pt, left=5pt, right=5pt, top=3pt, bottom=3pt,
                    title={\scriptsize\textbf{Tool result (returned by MCP server) \textbullet\ 5 content blocks}},
                    coltitle=white, colbacktitle=yellow!60!black,
                    fonttitle=\bfseries]
\scriptsize\ttfamily
\textbf{[1] TextContent:}\\
\quad\# introduction-to-styles/applying-styles/guide.md\\[1pt]
\quad Apply a paragraph style via the Paragraph Style dropdown at the\\
\quad far left of the Formatting toolbar. The widget is a combo box:\\
\quad a text field on the left and a separate dropdown arrow on the\\
\quad right. Click the arrow to open the style menu. See `fig01.png`.\\[3pt]
\textbf{[2] ImageContent} (\textit{paragraph-style menu open})\textbf{:}\\
\quad mime=image/png, data=$<$base64 PNG bytes, $\sim$140\,KB$>$\\[3pt]
\textbf{[3] TextContent:}\\
\quad Pick the style by clicking its row. The dropdown closes and the\\
\quad text field updates to the chosen style name; the status bar\\
\quad updates to the corresponding outline level. See `fig02.png`.\\[3pt]
\textbf{[4] ImageContent} (\textit{``Heading~1'' applied, status bar updated})\textbf{:}\\
\quad mime=image/png, data=$<$base64 PNG bytes, $\sim$120\,KB$>$\\[3pt]
\textbf{[5] TextContent:}\\
\quad Keyboard shortcuts: Ctrl+1..5 apply Heading~1..5 to the current\\
\quad paragraph; Ctrl+0 applies Default Paragraph Style. \dots
  \end{tcolorbox}

  \begin{tcolorbox}[colback=cyan!4, colframe=cyan!50!black, sharp corners,
                    boxrule=0.5pt, left=5pt, right=5pt, top=3pt, bottom=3pt,
                    title={\scriptsize\textbf{Turn $k{+}1$ \textbullet\ Agent (assistant message + next tool call)}},
                    coltitle=white, colbacktitle=cyan!50!black,
                    fonttitle=\bfseries]
\scriptsize\ttfamily
The guide says the dropdown has two click targets: a text field on the\\
left and a separate arrow on the right. Clicking the arrow opens the\\
menu (block~[2]). I'll click the arrow.\\[3pt]
computer\_use(action="click", x=196, y=156)~~~~\# the dropdown arrow
  \end{tcolorbox}
  \caption{One full agent turn end-to-end. The agent (i)~scans the \texttt{SKILL.md} index in its system prompt and matches its current sub-goal against a \emph{when to use} line; (ii)~issues \texttt{load\_topic} for the chosen topic; (iii)~receives a single tool result whose content list interleaves \texttt{TextContent} and \texttt{ImageContent} blocks in the order figures are referenced in the guide; (iv)~consumes the interleaved guide on its next step, citing a specific image block when choosing the click target. Image payloads are shown as size annotations; everything else is verbatim from the transcript.}
  \label{fig:apil-transcript}
\end{figure*}

\paragraph{Notes on the loop.}
\begin{itemize}[leftmargin=*,topsep=2pt,itemsep=0pt]
  \item \textbf{Many calls per trajectory.} The agent may invoke \texttt{load\_topic} multiple times across a rollout (Rule~2 of the preface), typically as the task moves between UI surfaces (e.g.\ page layout, tables, styles).
  \item \textbf{Text and image blocks are first-class context.} Both block types in the returned content list count as ordinary multimodal context for the agent's next assistant turn; no separate ``image attachment'' channel is used. This is why the multimodal vs.\ text-only contrast at evaluation time isolates the modality of presentation rather than any difference in how the content reaches the model.
  \item \textbf{No \texttt{Read} fallback for figures.} Rule~3 of the preface explicitly forbids issuing \texttt{Read} calls for \texttt{figXX.png} files in the skill. This is enforced socially (by the prompt) rather than mechanically; in practice agents comply because \texttt{load\_topic} already delivers the figures.
\end{itemize}

\clearpage
%
%
%

\graphicspath{{sections/appendix/appendix_training_supervision/figures/}{appendix/appendix_training_supervision/figures/}{appendix_training_supervision/figures/}{sections/appendix/appendix_skill_overview/figures/}{figures/}{./}}

\section{Three Construction-Stage Examples}
\label{app:examples}

This appendix grounds the three sub-pipelines of Section~\ref{sec:method-construct} in concrete artefacts. Section~\ref{app:examples-stage1} walks through one Stage~1 topic, end-to-end, from a page of the official LibreOffice Writer manual to the generated multimodal \texttt{guide.md}. Section~\ref{app:examples-stage2a} traces one region of the Stage~2(a) free UI explorer, from the planner's segmentation of the idle window to a worker's per-control notes and figures. Section~\ref{app:examples-stage2b} traces one target of the Stage~2(b) trajectory-targeted explorer, from a failed training rollout that surfaced a grounding gap, through the reviewer's diagnosis, to the worker capture and the patch landing in the matching topic.

All three examples are drawn from the actual Writer-skill build artefacts; screenshots, action traces, and quoted JSON/Markdown are verbatim from those artefacts unless explicitly noted as paraphrase.

\subsection{Stage 1: From Authored Documentation}
\label{app:examples-stage1}

We trace topic $t = $ \texttt{formatting-text/character-formatting} through the four LLM calls that make up Stage~1.

\paragraph{Step 1 \textbullet\ Source material.}
The input to Stage~1 is the official \textit{Getting Started with LibreOffice} guide for Writer 7.3, distributed as a PDF $D$. The chapter on character formatting spans pages 84--92 of the PDF, contains body text, and ships with two vendor-drawn screenshots: a view of the Character section of the Sidebar's Properties deck, and a view of the full \textbf{Format $>$ Character\dots} dialog. These two figures become the elements of $F_t$ for this topic; they are extracted unchanged from the PDF.

\paragraph{Step 2 \textbullet\ Topic entry from the TOC parser.}
The TOC parser runs once for the whole document and emits the topic set $\mathcal{T}$ as JSON. The entry for our topic is reproduced verbatim below.

\begin{tcolorbox}[colback=gray!4, colframe=gray!50, sharp corners,
                  boxrule=0.5pt, left=6pt, right=6pt, top=4pt, bottom=4pt,
                  fontupper=\ttfamily\scriptsize]
\{\\
\quad "topic\_id"  : "formatting-text/character-formatting",\\
\quad "title"     : "Formatting Characters",\\
\quad "parent\_id" : "formatting-text",\\
\quad "page\_start": 84,\\
\quad "page\_end"  : 92\\
\}
\end{tcolorbox}

\begin{figure}[h]
  \centering
  \begin{tcolorbox}[colback=cyan!4, colframe=cyan!50!black, sharp corners,
                    boxrule=0.6pt, left=5pt, right=5pt, top=4pt, bottom=4pt,
                    title={\scriptsize\textbf{Stage~1 output \textbullet\ \texttt{formatting-text/character-formatting/guide.md} (excerpt)}},
                    coltitle=white, colbacktitle=cyan!50!black,
                    fonttitle=\bfseries]
\scriptsize
\textbf{\# Formatting Characters}\\[2pt]
The quickest way to format characters is with \textbf{character styles} --- select your text, then pick a style from the Character Styles tab on the Styles deck in the Sidebar, the \textbf{Styles} menu, or the Formatting (Styles) toolbar.\\[2pt]
Direct formatting is fine for one-off tweaks: select your text and use the buttons on the Formatting toolbar or the Character panel in the Properties deck of the Sidebar. You'll find font name, size, bold, italic, underline, strikethrough, font color, highlight color, superscript, subscript, and more.\\[2pt]
\textit{See} \texttt{`fig01.png`}.\\[6pt]
\begin{center}
\fbox{\includegraphics[width=0.5\linewidth]{fig01}}\\[1pt]
{\tiny\textcolor{gray}{fig01.png \textbullet\ vendor screenshot extracted from PDF page 85}}
\end{center}
~\\[-4pt]
For the full set of options, open \textbf{Format $>$ Character\dots} The dialog has six tabs: Font, Font Effects, Position, Hyperlink, Highlighting, Borders.\\[2pt]
\textit{See} \texttt{`fig02.png`}.\\[6pt]
\begin{center}
\fbox{\includegraphics[width=0.5\linewidth]{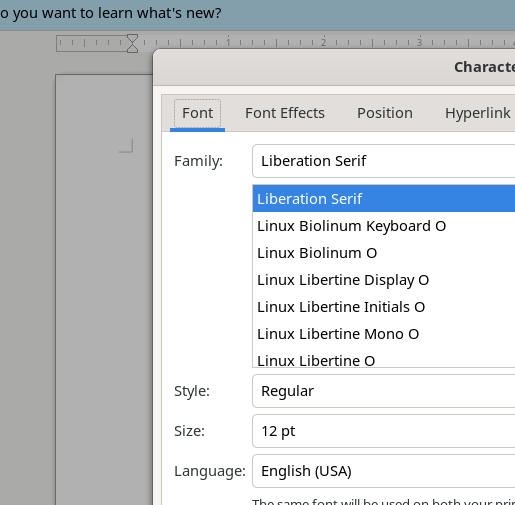}}\\[1pt]
{\tiny\textcolor{gray}{fig02.png \textbullet\ vendor screenshot extracted from PDF page 88}}
\end{center}
  \end{tcolorbox}
  \caption{Stage~1 output for one topic. The text body $p_t^{\mathrm{mm}}$ is generated from the concatenated page text of pages 84--92 of the official manual; the two figures are extracted unchanged from the PDF. The text-only body $p_t^{\mathrm{txt}}$ is generated in the same LLM call (joint generation, Section~\ref{sec:method-def}) and absorbs the visual content of \texttt{fig01}/\texttt{fig02} into prose; we omit it here and show only the multimodal variant.}
  \label{fig:apex-stage1-guide}
\end{figure}

\paragraph{Step 3 \textbullet\ Per-topic guide generation.}
The joint-generation call (Section~\ref{sec:method-def}) is invoked once for this topic with three inputs: the topic record (above), the concatenated text of pages 84--92, and the two extracted figures attached as image content. Claude Opus produces the multimodal text body $p_t^{\mathrm{mm}}$. Figure~\ref{fig:apex-stage1-guide} reproduces the opening of the generated \texttt{guide.md} alongside the vendor figures it references.

\paragraph{Step 4 \textbullet\ Index entry generation.}
A separate one-shot LLM call generates the one-sentence \emph{when to use} criterion that the agent will match against its current task at inference (Section~\ref{sec:method-loading}). The generated line for this topic is reproduced verbatim below; this is the line the agent sees in \texttt{SKILL.md} (Appendix~\ref{app:skill-overview-structure}).

\begin{tcolorbox}[colback=gray!4, colframe=gray!50, sharp corners,
                  boxrule=0.5pt, left=6pt, right=6pt, top=4pt, bottom=4pt,
                  fontupper=\ttfamily\scriptsize]
- [Formatting Characters](formatting-text/character-formatting/guide.md) -- Applying character styles and direct character formatting including fonts, size, color, and effects\\
\quad **Use when:** changing font name/size/style, applying character styles, adjusting character spacing and kerning, inserting hyperlinks via Character dialog, highlighting text, clearing direct formatting
\end{tcolorbox}

\paragraph{Notes on Stage 1.}
\begin{itemize}[leftmargin=*,topsep=2pt,itemsep=0pt]
  \item \textbf{Figures are vendor-drawn.} $F_t$ at Stage~1 contains screenshots authored by LibreOffice's documentation team. They show idealised, marketing-clean UI states (no real document content) and capture the application as the documentation was last revised; they may lag the shipped UI by one or more minor versions. Section~\ref{app:examples-stage2b} shows what happens at test time when that lag matters.
  \item \textbf{Coverage is whatever the docs cover.} Topics for which the manual lacks a section --- e.g.\ context menus added in a maintenance release, or recently restyled dialogs --- have no Stage~1 entry and rely on Stage~2 to enter the skill at all.
  \item \textbf{Matched-pair invariants.} The same joint-generation call produces $p_t^{\mathrm{mm}}$ and $p_t^{\mathrm{txt}}$ from the same context; the shared \texttt{SKILL.md} index entry is generated once and is byte-identical between the two skills.
\end{itemize}

\subsection{Stage 2(a): Free UI Explorer}
\label{app:examples-stage2a}

We trace one region of the Stage~2(a) free explorer for the Writer skill, target \texttt{character\_position\_dialog}.

\paragraph{Step 1 \textbullet\ Planner segments the idle window.}
The Stage~2(a) planner agent $P$ (Opus-class, Section~\ref{sec:method-construct}) is launched with a single screenshot of the freshly opened Writer application and instructed to partition the visible UI into $K{=}8$ regions covering everything the user can interact with. The planner's output is a JSON list of region records; the entry for the region we will trace is reproduced below (paraphrased for brevity; the full record also lists adjacent toolbars).

\begin{tcolorbox}[colback=gray!4, colframe=gray!50, sharp corners,
                  boxrule=0.5pt, left=6pt, right=6pt, top=4pt, bottom=4pt,
                  fontupper=\ttfamily\scriptsize]
\{\\
\quad "region\_id": "character\_position\_dialog",\\
\quad "name"     : "Format > Character dialog, Position tab",\\
\quad "entry"    : "Format menu > Character\dots > Position tab",\\
\quad "scope"    : "Document the Position tab of the Character dialog: \\
\qquad Superscript / Subscript / Normal radios, Raise/lower by, \\
\qquad Relative font size, Rotation/Scaling, Spacing, Pair kerning."\\
\}
\end{tcolorbox}

\paragraph{Step 2 \textbullet\ Worker drives the region.}
A Sonnet-class worker $W_i$ is spawned in a clean Writer Docker container with the region record above as its prompt. The worker opens the dialog, clicks each control, captures cropped screenshots of state changes, and writes a structured note. Two of the worker's captures are reproduced below; the worker emitted twelve in total across all controls in the region.

\begin{figure}[h]
  \centering
  \begin{subfigure}[t]{0.48\columnwidth}
    \includegraphics[width=\linewidth]{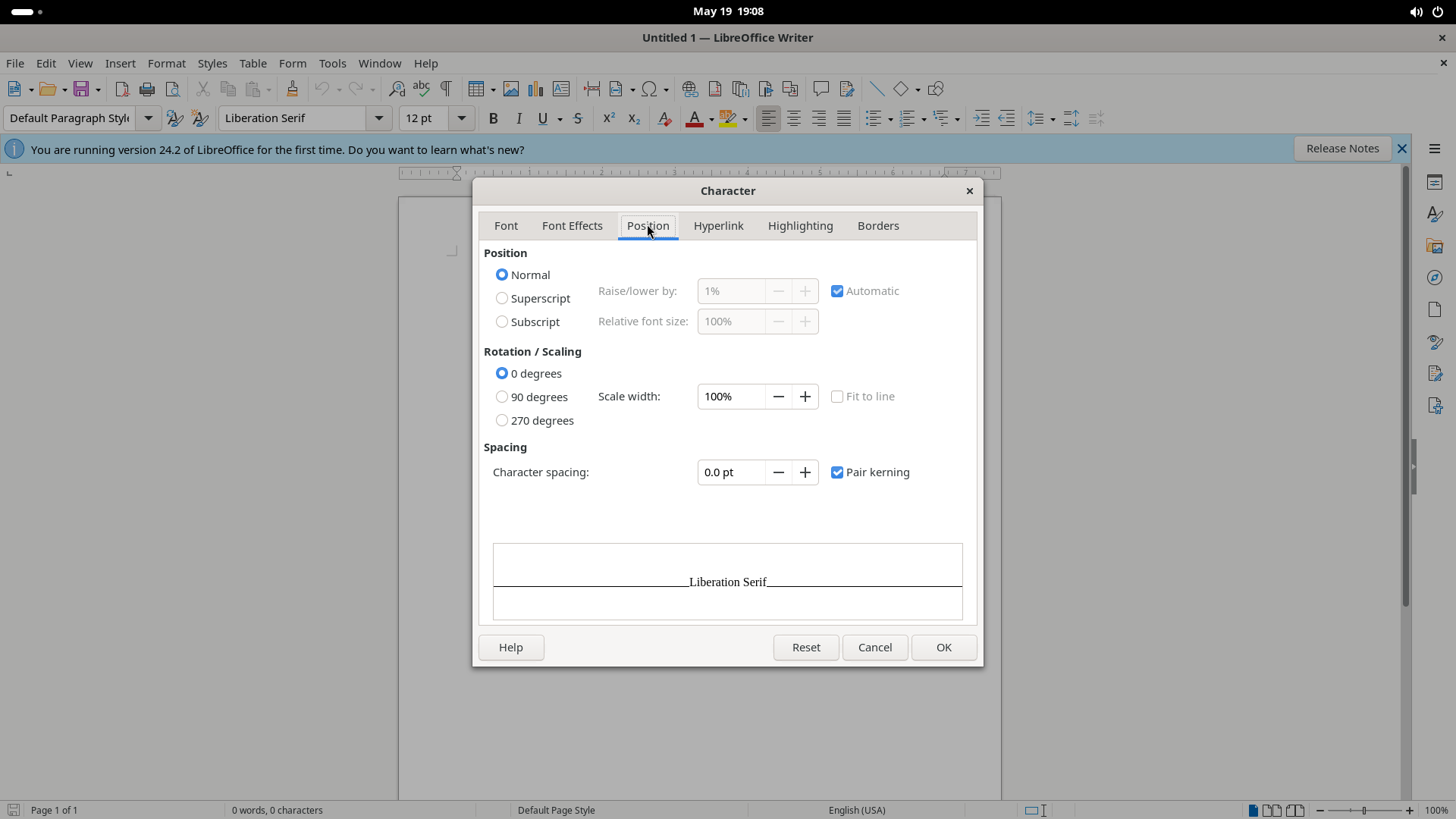}
    \caption{Position tab on first open: Normal radio selected, Raise/lower by and Relative font size greyed.}
    \label{fig:apex-s2a-cap1}
  \end{subfigure}\hfill
  \begin{subfigure}[t]{0.48\columnwidth}
    \includegraphics[width=\linewidth]{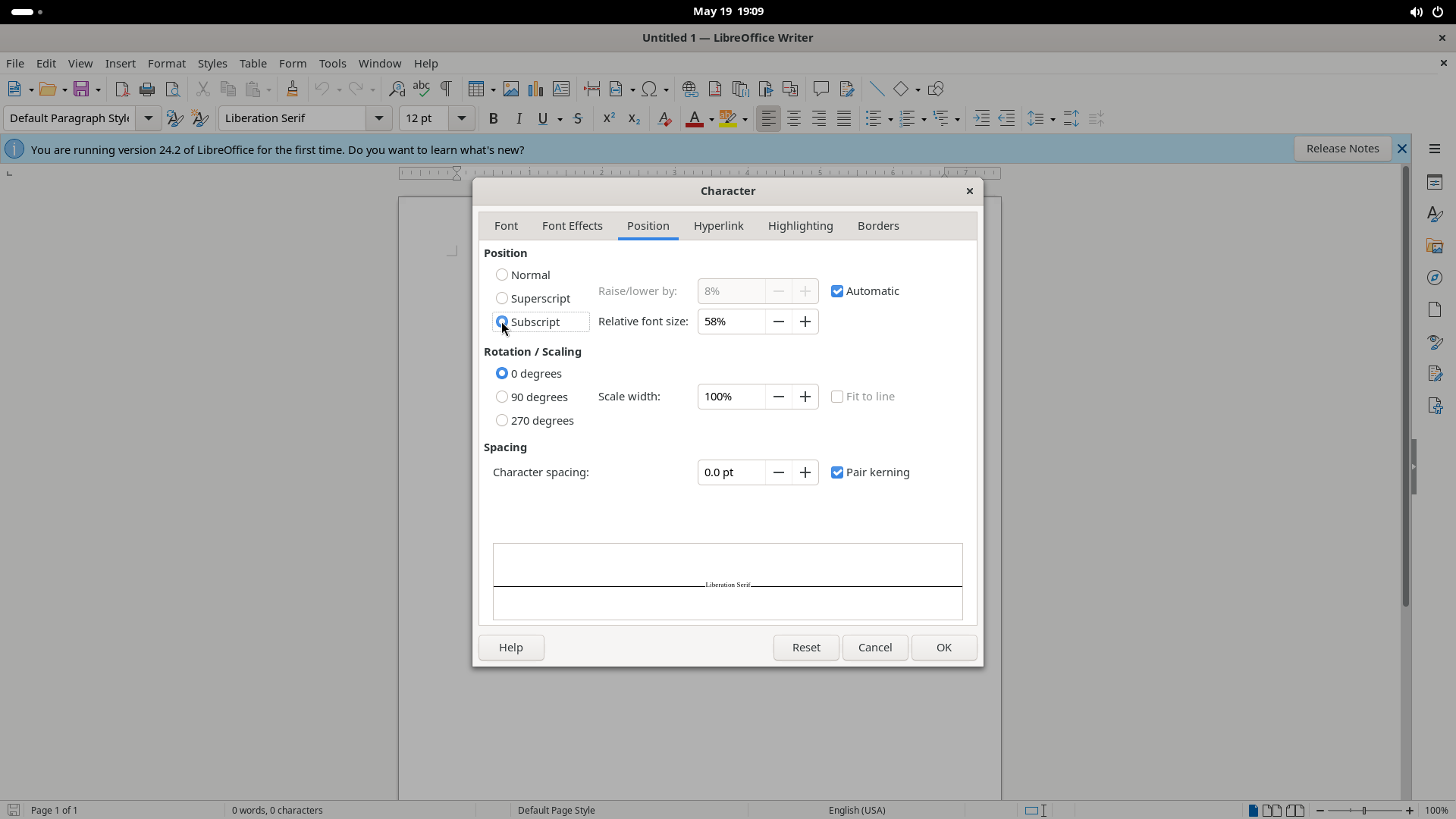}
    \caption{After clicking the Subscript radio: Raise/lower auto-set to 8\%, Relative font size to 58\%, Automatic checkbox auto-checked, preview pane shows subscripted text.}
    \label{fig:apex-s2a-cap2}
  \end{subfigure}
  \caption{Two of the worker's twelve captures for region \texttt{character\_position\_dialog}. The worker reaches each state by issuing real clicks in the live application; the screenshot in (b) is taken immediately after the Subscript click and shows the resulting auto-population of the spinner fields.}
  \label{fig:apex-s2a-captures}
\end{figure}

\paragraph{Step 3 \textbullet\ Per-control notes.}
Alongside the captures, the worker emits a structured Markdown note that names every control, its pixel-coordinate location, its state-transition behaviour, and which screenshot witnesses the behaviour. An excerpt for two of the radios is reproduced verbatim below (compressed line wrapping is the only edit).

\begin{tcolorbox}[colback=gray!4, colframe=gray!50, sharp corners,
                  boxrule=0.5pt, left=6pt, right=6pt, top=4pt, bottom=4pt,
                  fontupper=\ttfamily\scriptsize]
\#\#\# 3. Superscript radio button\\
- Location: Second radio in the Position group; approx (662, 393) in 1920x1080\\
- Behaviour: Activates superscript positioning. Sets Raise/lower by to 33\% (upward) and Relative font size to 58\% as automatic defaults. Raise/lower spinner becomes read-only when Automatic is checked.\\
- Evidence: step\_005.png\\
~\\
\#\#\# 4. Subscript radio button\\
- Location: Third radio in the Position group; approx (662, 423) in 1920x1080\\
- Behaviour: Activates subscript positioning. Sets Raise/lower by to 8\% (downward) and Relative font size to 58\% as automatic defaults.\\
- Evidence: step\_006.png
\end{tcolorbox}

\paragraph{Step 4 \textbullet\ Assembly into a topic.}
After all $K$ workers return, the assembler agent consolidates each region's notes and figures into a per-region reference section, and an LLM mapper decides which Stage~1 topic each region most directly enriches. For this region, the mapper points at \texttt{formatting-text/} \texttt{character-formatting/} (the same topic Stage~1 generated from pages 84--92 of the manual, Section~\ref{app:examples-stage1}). The assembled section is appended to that topic's \texttt{guide.md} and the worker's reference screenshot (Figure~\ref{fig:apex-s2a-captures}\textcolor{red!70!black}{b} re-shot at full resolution) is added to $F_t^{\mathrm{mm}}$.

\paragraph{Why ``free'' matters.}
The free explorer reaches UI surfaces the manual is silent on. The Position tab \emph{is} documented in the official guide, but with a vendor-drawn screenshot of the dialog at its default state; the manual does not show that selecting Subscript auto-populates the spinner fields, nor that the Automatic checkbox auto-checks. Those state-transition facts come from the worker's free-exploration captures and are what Section~\ref{app:examples-stage2b} shows are decisive at agent test time.

\subsection{Stage 2(b): Trajectory-Targeted Explorer}
\label{app:examples-stage2b}
\label{app:trajectory-targeted-example}

The third example traces one target of the Stage~2(b) trajectory-targeted explorer, from a failed training rollout that surfaced a UI grounding gap, through the reviewer agent's diagnosis, to the worker capture and the patch landing in the matching topic. The worked example is target \texttt{paragraph\_style\_applicator}, the Paragraph Style dropdown in LibreOffice Writer's formatting toolbar.

\paragraph{Step 1 \textbullet\ Training task.}
\texttt{exam\_paper\_formatting} is one of the 16 training tasks in the Writer training split. It asks the agent to reformat a biology exam paper end-to-end: set margins, font, line spacing, format the title block, italicise the \textsc{instructions} paragraph, apply the Heading~2 paragraph style to the three section headers (``Part~A'', ``Part~B'', ``Part~C''), and save the result as \texttt{bio101\_final\_exam.docx}.

\begin{figure}[h]
  \centering
  \includegraphics[width=\columnwidth]{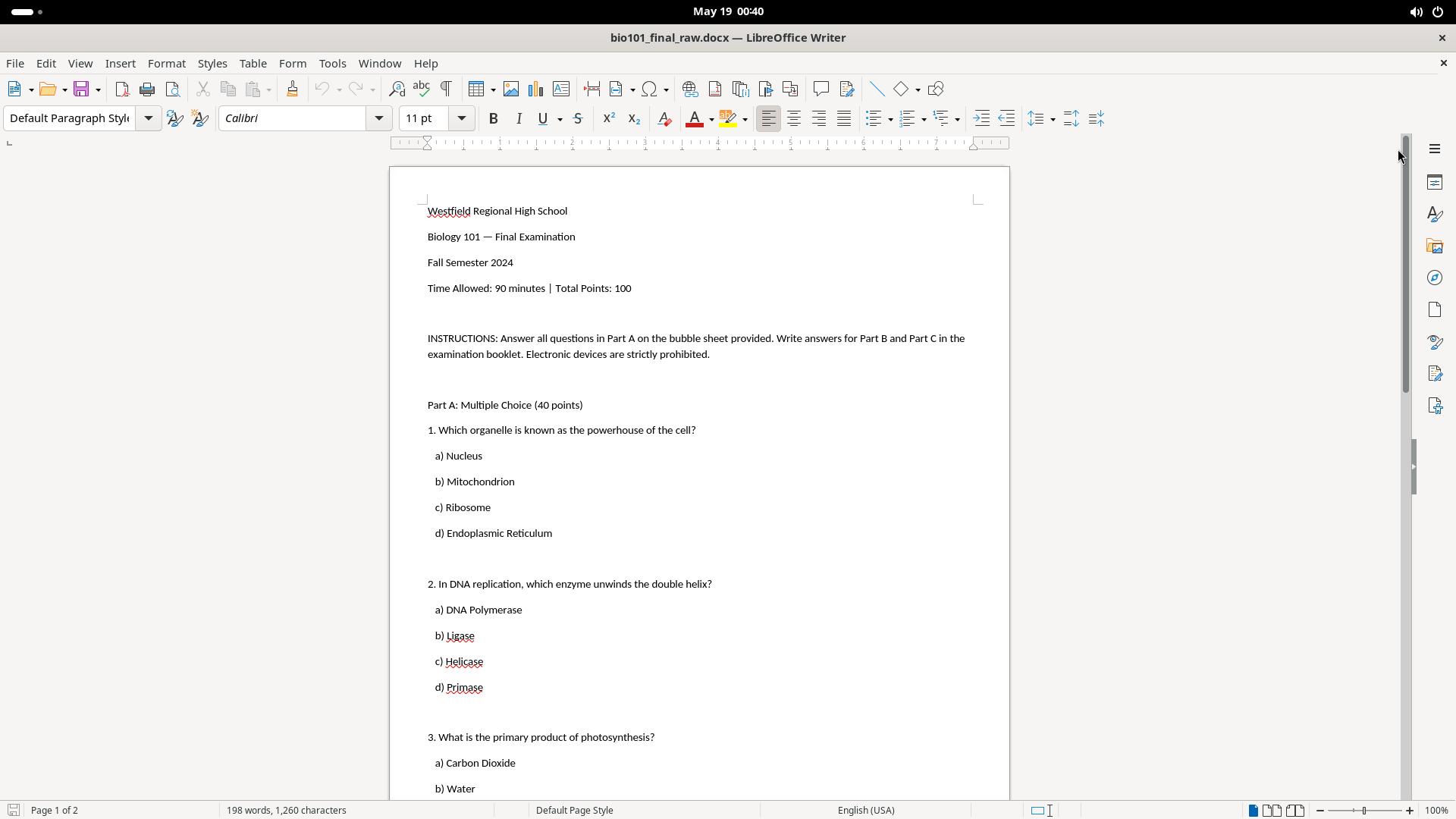}
  \caption{Initial state of \texttt{exam\_paper\_formatting}. The three target headers (``Part~A:~Multiple Choice (40 points)'' and the two parallel headers below) are plain body text. The paragraph-style dropdown in the top-left of the formatting toolbar reads \emph{``Default Paragraph Style''}. The sub-goal is to change those three headers to \emph{``Heading 2''}.}
  \label{fig:apex-s2b-initial}
\end{figure}

This subsection focuses on the Heading~2 sub-goal, which is the part that triggered the reviewer agent's diagnosis. Figure~\ref{fig:apex-s2b-initial} shows the document's initial state.

\noindent\textbf{Outcome with the Stage~1 multimodal skill.}
\begin{itemize}[leftmargin=*,topsep=2pt,itemsep=0pt]
  \item Verifier score: $0.000$ (FAIL).
  \item Steps used: $80\,/\,80$ (budget exhausted).
  \item Verifier feedback: \emph{``Output file 'bio101\_final\_exam.docx' not found.''}
\end{itemize}
The agent applied margins, font, line spacing, title formatting and italicisation successfully, and eventually got Heading~2 applied to each section header as well. It ran out of budget before reaching the final Save~As. The next step explains why.

\paragraph{Step 2 \textbullet\ Where the agent was inefficient.}
The paragraph-style dropdown is actually \emph{two} adjacent widgets glued into one combo box: a text field (left, $\sim$\,(75,118) in 1920$\times$1080) and a separate $\blacktriangledown$ arrow (right, $\sim$\,(196,156)). Clicking the $\blacktriangledown$ arrow opens a menu listing every style (Heading~1, Heading~2, \dots); clicking the desired style applies it. The whole interaction is two clicks, no typing, no Enter.

The Stage~1 skill's prose mentions the dropdown but provides no figure of the open menu and no callout that the arrow is a distinct click target. The agent ends up clicking the text field at (75, 118), then tries to type-then-Enter its way to the style instead. Table~\ref{tab:apex-s2b-trace} shows the verbatim 8-action trace (rollout indices 141--153).

\begin{table}[h]
  \centering
  \scriptsize
  \setlength{\tabcolsep}{4pt}
  \renewcommand{\arraystretch}{1.15}
  \begin{tabular}{@{}r l >{\raggedright\arraybackslash}p{0.45\columnwidth}@{}}
    \toprule
    \# & Action & Comment \\
    \midrule
    1 & \texttt{click(75, 118)}         & clicked the text field, not the $\blacktriangledown$ arrow \\
    2 & \texttt{triple\_click(...)}     & \textcolor{red!70!black}{wasted} \\
    3 & \texttt{triple\_click(...)}     & \textcolor{red!70!black}{wasted} \\
    4 & \texttt{hotkey ctrl+a}          & fall back to Ctrl+A \\
    5 & \texttt{type\_text "Heading 2"} & replaces highlighted text \\
    6 & (verify screenshot)             & \textcolor{red!70!black}{wasted} \\
    7 & \texttt{key\_press Return}      & this is what actually applies \\
    8 & (verify screenshot)             & \textcolor{red!70!black}{wasted} \\
    \bottomrule
  \end{tabular}
  \caption{Verbatim 8-action trace for one Heading~2 application (rollout indices 141--153). The canonical $\blacktriangledown$-then-pick path is only 2 actions.}
  \label{tab:apex-s2b-trace}
\end{table}

\noindent\textbf{What the agent did not understand:} that the dropdown is composed of two click targets, not one. It treated the entire box as a text field, picked the easy click target (the text-field portion), and never tried the $\blacktriangledown$ arrow that would have opened the menu of style choices. The Stage~1 prose said ``use the dropdown to apply a style'' without showing the open-menu state or specifying which click target opens it. The cost: $\sim$6 wasted actions per heading $\times$ 3 section headers $\approx$ 18 wasted actions, against an 80-step budget that the agent then runs out of before the closing Save~As.

\paragraph{Step 3 \textbullet\ Reviewer agent identifies the weak UI region.}
The reviewer agent $V$ (Section~\ref{sec:method-construct}, Stage~2(b)) reads all 16 training rollouts and looks for one signal: the same widget breaks multiple agents in similar ways. Four corroborating failures appear for the paragraph-style dropdown:
\begin{itemize}[leftmargin=*,topsep=2pt,itemsep=0pt]
  \item \texttt{exam\_paper\_formatting}: 8-action fumble per heading, ran out of budget (this subsection).
  \item \texttt{format\_research\_paper}: used \texttt{Ctrl+1}/\texttt{Ctrl+2} hotkeys that did not persist as Heading styles (0/7 Heading~1, 0/4 Heading~2).
  \item \texttt{clinical\_protocol\_formatting}: 0/9 major headings styled.
  \item \texttt{outline\_numbering\_handbook}: same dropdown gap.
\end{itemize}
Four independent failures on the same widget, in different ways, rule out random budget exhaustion as the sole cause and corroborate a systematic UI gap. The reviewer writes one target:

\begin{tcolorbox}[colback=gray!4, colframe=gray!50, sharp corners,
                  boxrule=0.5pt, left=6pt, right=6pt, top=4pt, bottom=4pt,
                  fontupper=\small]
\textbf{target\_id:} \texttt{paragraph\_style\_applicator}\\
\textbf{name:} Paragraph Style dropdown in the Formatting Toolbar\\
\textbf{scope:} Document the paragraph style dropdown (leftmost box in the formatting toolbar). \emph{Show how to click it, type a style name (e.g.~`Heading 1'), and press Enter to apply.} Cover the dropdown list behaviour, autocomplete, and what happens when an invalid name is typed.
\end{tcolorbox}

\begin{figure}[h]
  \centering
  \includegraphics[width=\columnwidth]{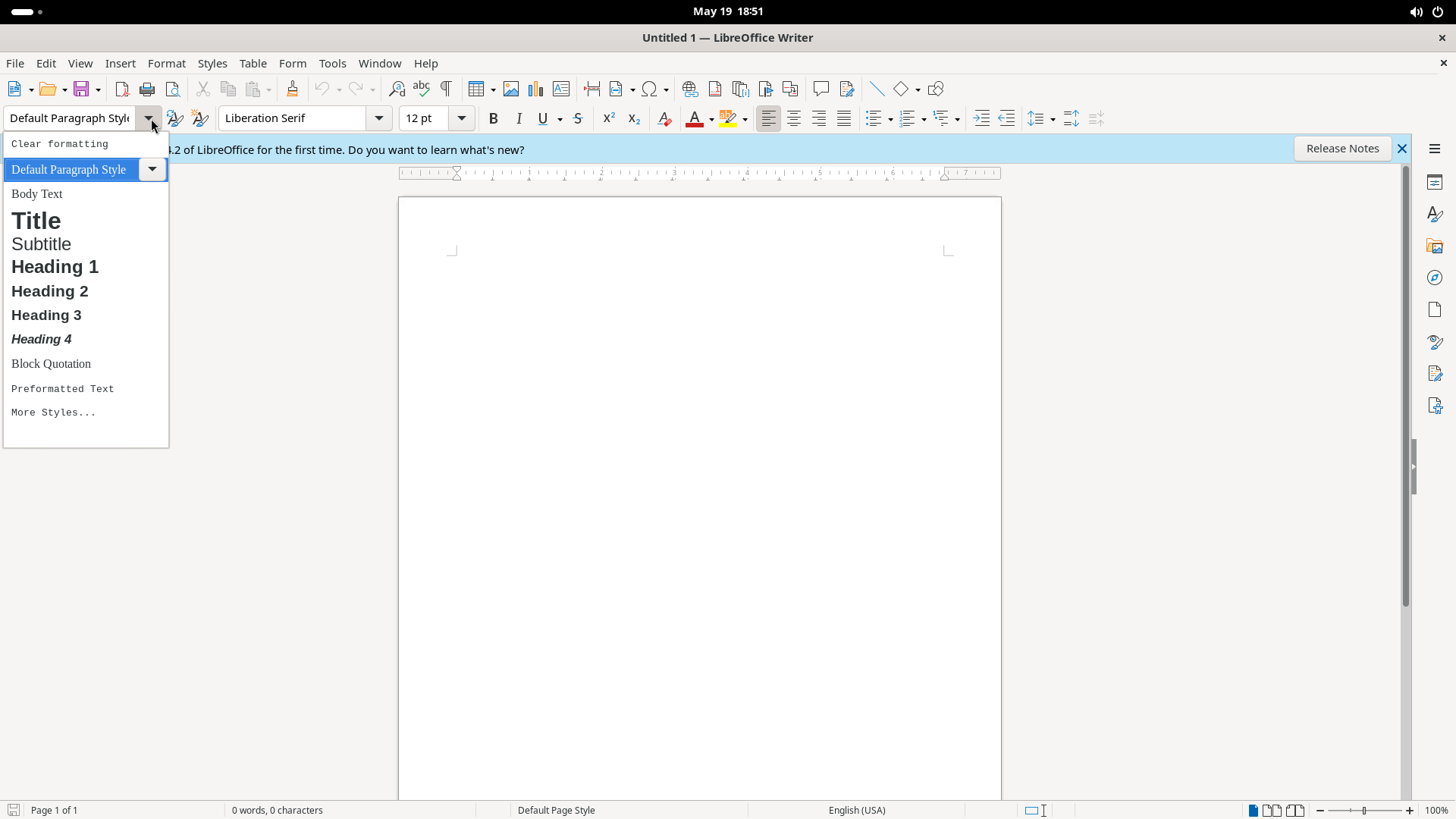}
  \caption{Worker capture, step~4: the $\blacktriangledown$ arrow has been clicked and the style menu is open. Every paragraph style is listed and rendered in its own font (Title large, Heading~1 large bold, Heading~2 large, \dots). This is the figure that lands in the patched skill as the primary reference.}
  \label{fig:apex-s2b-worker-menu}
\end{figure}

\begin{figure}[h]
  \centering
  \includegraphics[width=\columnwidth]{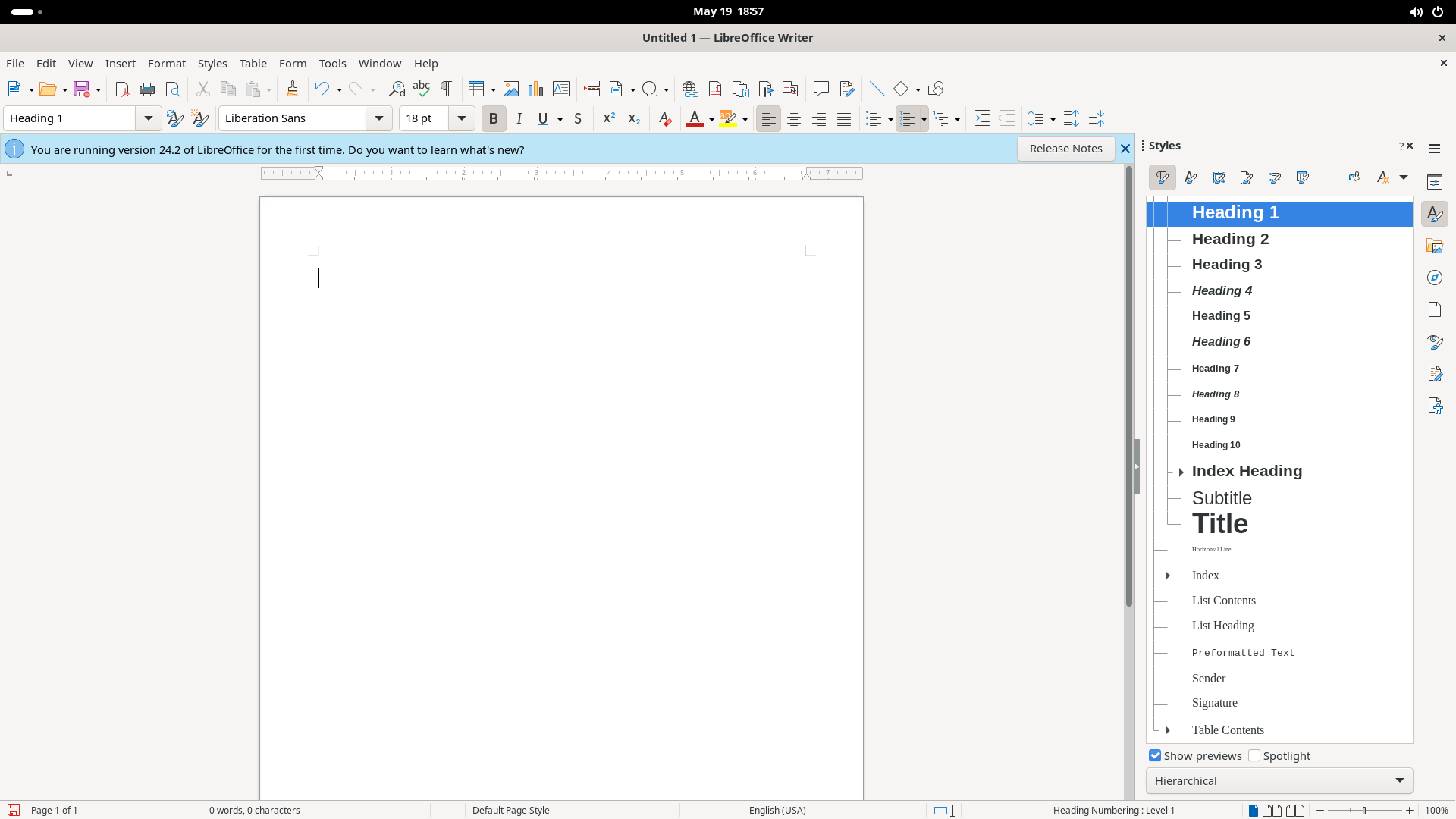}
  \caption{Worker capture, step~38: after selecting ``Heading 1'' from the menu, the dropdown text field updates to ``Heading 1'' (top-left) and the status bar reads ``Heading Numbering : Level 1'' (bottom-right). This is the canonical applied-state confirmation a future agent should recognise.}
  \label{fig:apex-s2b-worker-applied}
\end{figure}

\paragraph{Step 4 \textbullet\ Trajectory-targeted worker captures the menu.}
A worker $W'_i$ is spawned in a clean Writer container with the above target as its prompt. The worker uses the same protocol as the Stage~2(a) free workers (Section~\ref{app:examples-stage2a}): real clicks in the live application, cropped screenshots of state changes, per-control notes. It clicks the $\blacktriangledown$ arrow, captures the open menu (Figure~\ref{fig:apex-s2b-worker-menu}), then applies a style and captures the applied state (Figure~\ref{fig:apex-s2b-worker-applied}).

\paragraph{Step 5 \textbullet\ Patch lands and pays off at test time.}
The same assembler and LLM mapper used by Stage~2(a) (Section~\ref{app:examples-stage2a}) consolidate the worker's notes and figures and assign them to the \texttt{applying-styles} topic --- the topic the failing rollouts had loaded. The assembler appends a short paragraph from the worker's notes that names the two click targets (the text field at $\sim$(75,118), the $\blacktriangledown$ arrow at $\sim$(196,156)) and prescribes the canonical 2-click interaction, plus the two figures of menu-open and applied states. At inference, any task touching paragraph styles loads the patched topic via \texttt{load\_topic} (Appendix~\ref{app:inference-loading}) and receives both figures plus the 2-click recipe in one tool result. For \texttt{exam\_paper\_formatting} specifically, the freed $\sim$18 actions are exactly the budget the agent needed for the closing Save~As; and because the patch is scoped to the \emph{UI region} rather than to any one task, it transfers to every future test task that uses heading styles --- including the three other training-task failures the reviewer cited.

\clearpage

\section{Agent Setup}
\label{app:agent-details}

This appendix details the agent setup, the GUI tool surface, and the sandboxing setup deferred from Section~\ref{sec:exp-setup}.

\paragraph{Agent.}
The evaluated agent (Claude Opus~4.6) runs inside the \textbf{Claude Code CLI}, an off-the-shelf agentic harness that exposes MCP tool calls and standard read/write tools to the model. The same CLI image is reused across all skill conditions; only the skill folder mounted into the container changes. Per-step prompts are held fixed.

\paragraph{GUI tool surface.}
The agent interacts with the desktop through a single MCP tool, \texttt{computer} (from the open-source \texttt{computer-use-mcp} server), which exposes eleven actions; the full set is reproduced verbatim in Table~\ref{tab:computer-actions}. The action enum and the per-action behaviour are identical to those documented in the Anthropic computer-use reference. The agent obtains pixels by calling \texttt{get\_screenshot}; coordinates are reported in the API image space (downsampled to fit the 1.15\,MP / 1568\,px API limits) and scaled to logical screen pixels by the server, so the agent does not need to track the display resolution itself. Skill consultation is exposed by a \emph{separate} MCP server bundled inside each skill folder (\texttt{load\_topic} and \texttt{list\_topics}; Appendix~\ref{app:inference-loading}); the two MCP servers --- the GUI controller and the skill MCP server --- are independent and the agent sees both in its tool list.

\begin{table}[h]
\centering
\scriptsize
\setlength{\tabcolsep}{5pt}
\renewcommand{\arraystretch}{1.2}
\begin{tabular}{@{}l >{\raggedright\arraybackslash}p{0.60\columnwidth}@{}}
\toprule
\textbf{Action} & \textbf{Behaviour} \\
\midrule
\texttt{key}                  & Press a key or key combination. \\
\texttt{type}                 & Type a string of text. \\
\texttt{mouse\_move}          & Move the cursor to $(x, y)$. \\
\texttt{left\_click}          & Click the left button (optionally at $(x, y)$ first). \\
\texttt{left\_click\_drag}    & Drag from the current cursor to $(x, y)$. \\
\texttt{right\_click}         & Click the right button (optionally at $(x, y)$ first). \\
\texttt{middle\_click}        & Click the middle button. \\
\texttt{double\_click}        & Double-click the left button. \\
\texttt{scroll}               & Scroll at $(x, y)$ in direction \texttt{"up"} / \texttt{"down"} / \texttt{"left"} / \texttt{"right"}, optionally \texttt{":N"} pixels. \\
\texttt{get\_screenshot}      & Take a screenshot (returned as an image). \\
\texttt{get\_cursor\_position}& Return the current cursor $(x, y)$. \\
\bottomrule
\end{tabular}
\caption{Actions exposed by the \texttt{computer} MCP tool 
(\texttt{computer-use-mcp/src/tools/computer.ts}).}
\label{tab:computer-actions}
\end{table}

\paragraph{Two-container sandboxing.}
Each task runs in a \textbf{two-container} setup:
\begin{itemize}[leftmargin=*,topsep=2pt,itemsep=1pt]
  \item A \textbf{desktop-environment container} (Xvfb + the target application: LibreOffice, GIMP, OpenToonz, etc.) ships with the benchmark and provides the actual UI to drive. The benchmark's verifier and \texttt{init.max\_steps} budget come from this container.
  \item A \textbf{Claude Code CLI container} runs the agent, mounts both the per-task workspace and the skill folder read-only, and connects to the desktop container over an HTTP bridge (\texttt{GA\_BRIDGE\_URL}). All \texttt{computer} actions are forwarded to the desktop container by the bridge; verifier scoring and \texttt{env.step()} bookkeeping happen there.
\end{itemize}
The split keeps the agent fully sandboxed from the host while preserving deterministic per-task state: the desktop container is reset to a clean snapshot before each task. Both containers are spun up by a single \texttt{run\_task.py} entry point per task, which also writes an MCP-config file that registers the two MCP servers (controller + skill loader) in the Claude CLI container.

\paragraph{Step budget.}
What counts against the per-task budget is the number of \texttt{computer}-tool actions that mutate the desktop --- click, drag, type, key, hotkey, scroll --- as recorded by \texttt{env.step()} on the desktop side. \texttt{get\_screenshot}, \texttt{get\_cursor\_position}, \texttt{Read}, \texttt{load\_topic}, and \texttt{list\_topics} are observation / lookup operations and are \emph{not} charged against the budget; this prevents skill consultation from competing with the actual GUI interaction. The per-task ceiling is whatever the benchmark ships in \texttt{task.json}'s \texttt{init.max\_steps} (median 60 in our Writer split; min 40, max 80).

\clearpage

\section{Datasets}
\label{app:dataset-details}

OSExpert-Eval~\citep{liu2026osexpert} is a curated benchmark designed to evaluate computer-use agents on professional-level challenges that go beyond routine GUI interaction. The benchmark consists of 113 tasks spanning three major categories: Long Horizon Compositional Workflows, Unseen UI Generalization, and Fine-Grained Action Execution. Long Horizon contains 30 tasks, including 24 Office tasks and 6 GIMP tasks, emphasizing multi-step workflows that require composing multiple unit functions in
a correct and robust order. Unseen UI contains 50 tasks, including 20 Tableau tasks and 30 MiniWord tasks, targeting novel layouts and interaction patterns that are uncommon
in current agents’ training distributions. Fine-Grained contains 33 tasks, including 14 GIMP tasks and 19 Office tasks, requiring precise low-level control such as accurate
text selection, object manipulation, and spatial alignment. Among these three categories, Long Horizon and Unseen UI are exactly the scenarios where agent skills are expected to be helpful. For the Fine-Grained category, Claude Code CLI powered by Claude Opus 4.6 already performs well, so we do not include this subset. Therefore, we include all tasks from the Long Horizon category, as well as the Tableau domain from Unseen UI. (The MiniWord domain in this category has missing verifier files, which is why we exclude it here.) The dataset is released under the MIT License.

CUA-World~\citep{aggarwal2026gym} is a large-scale benchmark consisting of over 10K long-horizon computer-use tasks spanning diverse domains, including medical science, astronomy, engineering, creative tools, and enterprise software applications. Each environment is configured with realistic data and includes predefined train/test splits. The environments and tasks are generated through an agentic pipeline with executable verifiers for evaluation. In this work, we manually verify and curate five domains for our experiments: LibreOffice Writer, LibreOffice Calc, LibreOffice Impress, QGIS, and OpenToonz. The dataset is released under the MIT License.

\section{Results of \textsc{Qwen3.5-397B-A17B-FP8}}
\label{app:qwen3}
To see the generalizability of our multimodal skill generation pipeline, we additionally evaluate \textsc{Qwen3.5-397B-A17B-FP8} on OSWorld. We selected the LibreOffice domain for skill reusing. As show in Table~\ref{tab:qwen3}, tasks with text-skill achieve the highest scores, while tasks with multimodal skills get relativly similar results as the no-skill baseline. Investigating as the trajectories, we induce the reason as the burdon of overly long context. Since when loading related topics from multimodal skills, all figures in that topic are dumped into the context. Such context burdon can lead to degradation in model behavior such as malformating in tool calling.

\definecolor{banner}{HTML}{DCE9F4}
\definecolor{altrow}{HTML}{F2F2F2}

\begin{table}[t]
\centering
\scriptsize
\setlength{\tabcolsep}{5pt}
\renewcommand{\arraystretch}{1.25}
\begin{tabular}{l cccc}
\toprule
& \multicolumn{3}{c}{\textbf{OSWorld (LibreOffice)}}
  & \\
\cmidrule(lr){2-4}
\textbf{Method}
  & Writer & Calc & Impress
  & \textbf{Avg} \\
  & {\scriptsize (23)} & {\scriptsize (47)} & {\scriptsize (47)}
  & \\
\midrule
No-skill baseline             & 0.652 & 0.553 & 0.467 & 0.538 \\
Text skill                    & 0.696 & 0.596 & 0.466 & 0.563 \\
\rowcolor{altrow}
Multimodal skill              & 0.652 & 0.532 & 0.466 & 0.529 \\
\bottomrule
\end{tabular}
\caption{Cross-model replication on \textbf{OSWorld} with \textbf{Qwen3-397B} as the underlying agent. Per-domain mean score in $[0,1]$; integer beneath each column header is the task count; \textbf{Avg} is the unweighted mean across the three LibreOffice domains. The two skill conditions are the Stage~1 (Documentation-only) variants in their text-matched and multimodal forms; UI-explorer (Stage~2) results are reported in Table~\ref{tab:main}.}
\label{tab:qwen3}
\end{table}

\section{Model and Compute Information}

Claude Code CLI was used as the primary agent for experiments. The approximate API usage cost during the project was around \$1,000.

We additionally used QWEN3.5-397B-A17B-FP8 for parts of the experimental pipeline.
Training and evaluation experiments were conducted on 8H100 GPUs for approximately 48 hours, corresponding to roughly 384 H100 GPU hours.

\clearpage

\section{Qualitative Analysis: Supporting Figures}
\label{app:qa-examples}

\begin{figure}[h]
  \centering
  \begin{subfigure}[t]{0.48\columnwidth}
    \includegraphics[width=\linewidth]{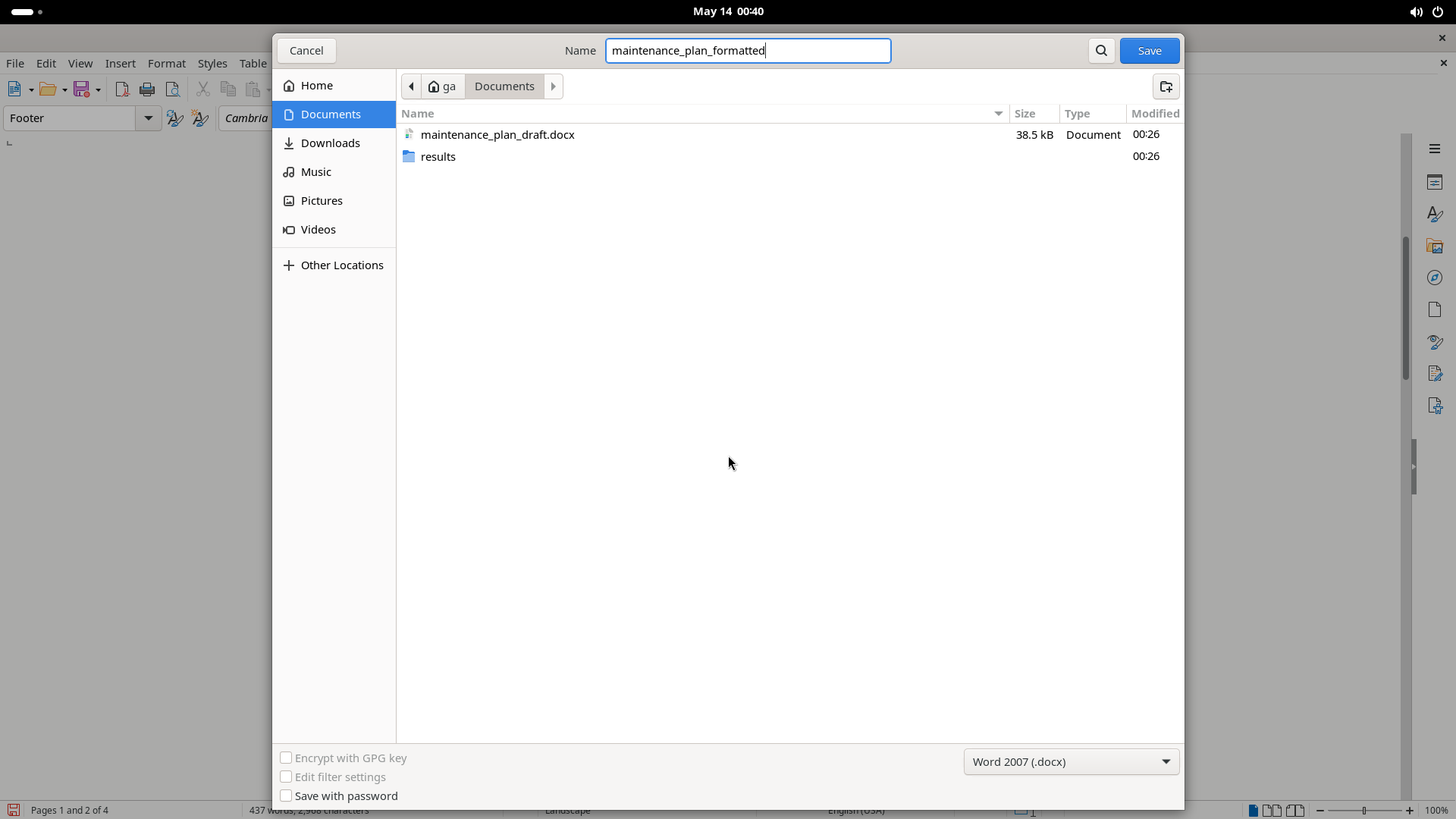}
    \caption{\textbf{Text-only skill.} Agent clicks \texttt{(1011, 44)} on the dialog title bar --- LibreOffice closes.}
    \label{fig:qa-ms-fail}
  \end{subfigure}\hfill
  \begin{subfigure}[t]{0.48\columnwidth}
    \includegraphics[width=\linewidth]{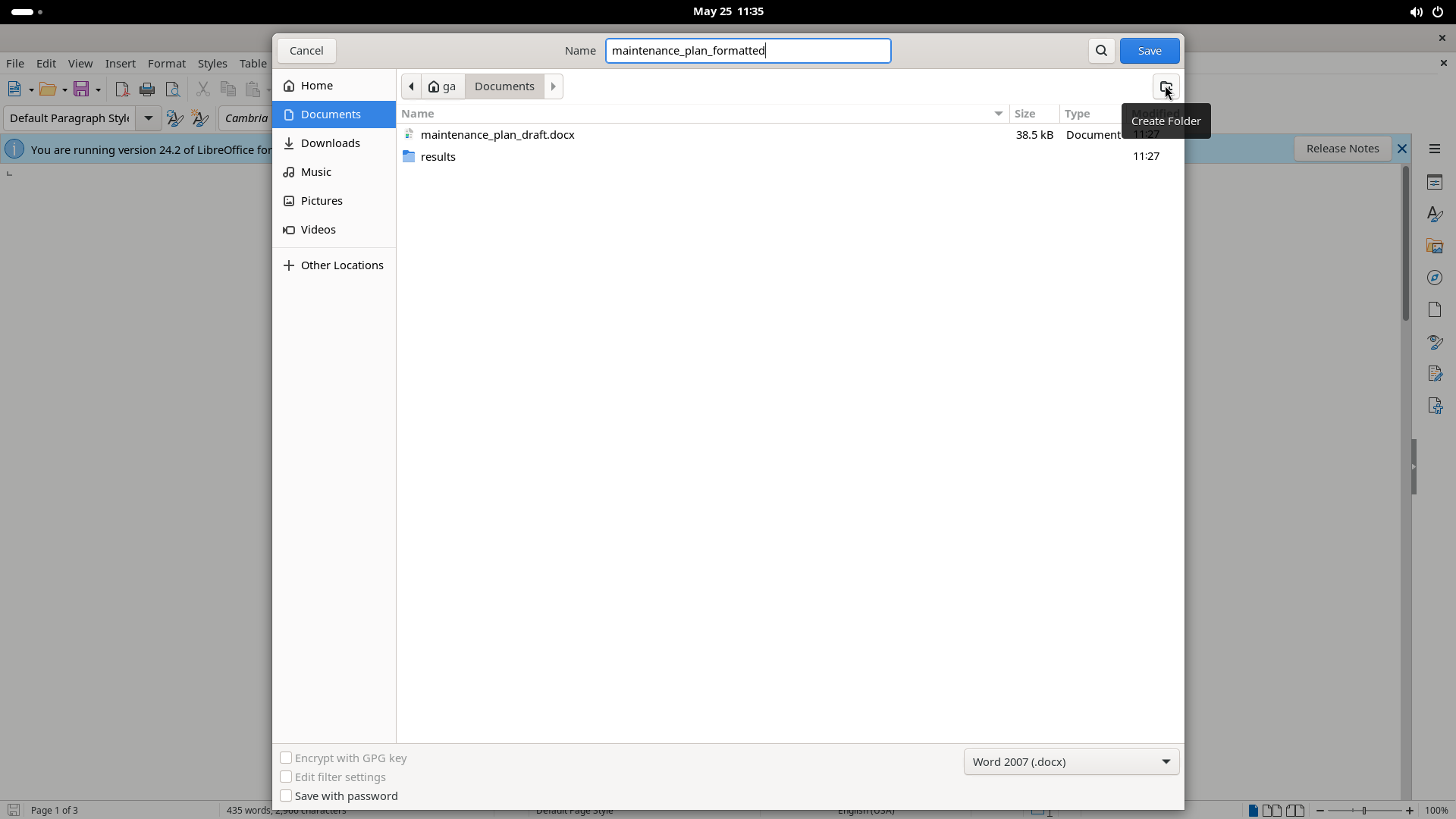}
    \caption{\textbf{Multimodal skill.} Reference screenshot delivered with the topic guide pins the correct Save button at the dialog footer.}
    \label{fig:qa-ms-worker}
  \end{subfigure}
  \caption{Case~(i), Save-button mis-location on test task \texttt{maintenance\_plan\_orientation}. Prose like ``Save at the bottom of the dialog'' is ambiguous; the reference screenshot removes the ambiguity.}
  \label{fig:qa-maintenance-save}
\end{figure}

This appendix provides supporting figures for the qualitative analysis in Section~\ref{sec:exp-qualitative}. Figure~\ref{fig:qa-maintenance-save} illustrates case~(i) on test task \texttt{maintenance\_plan\_orientation}: the text-only agent clicks the title bar of the Save~As dialog (closing LibreOffice), while the Stage~2(b) worker's captured screenshot, delivered to the multimodal agent through \texttt{load\_topic}, pins the Save button at its actual location in the dialog footer.

\end{document}